\documentclass[sn-mathphys,Numbered]{sn-jnl}
\usepackage[utf8]{inputenc} 

\usepackage[utf8]{inputenc}
\usepackage{graphicx}%
\usepackage{multirow}%
\usepackage{amsmath,amssymb,amsfonts}%
\usepackage{amsthm}%
\usepackage{mathrsfs}%
\usepackage[title]{appendix}%
\usepackage{textcomp}%
\usepackage{manyfoot}%
\usepackage{booktabs}%
\usepackage{algorithm}%
\usepackage{algorithmicx}%
\usepackage{algpseudocode}%
\usepackage{listings}%
\usepackage{glossaries}
\usepackage{ifthen}
\usepackage{subcaption}
\usepackage{multirow}
\usepackage[table,xcdraw]{xcolor}
\usepackage{adjustbox}
\usepackage{glossaries}
\usepackage{booktabs}
\usepackage{tabularx}

\usepackage{background}

\backgroundsetup{
  scale=6, 
  color=gray,
  opacity=0.3,
  angle=45, 
  position=current page.center, 
  vshift=0cm,
  hshift=0cm,
  contents={\textsf{Personal use only}}
}
\begin{document}
\hbadness=99999  
\newdimen\hfuzz

\title[Article Title]{Performance Comparison of ROS2 Middlewares for Multi-robot Mesh Networks in Planetary Exploration}


\author*[1]{Loïck Pierre Chovet}\email{loick.chovet@uni.lu}
\equalcont{These authors contributed equally to this work.}

\author[1]{Gabriel Manuel Garcia}\email{gabriel.garcia@uni.lu}
\equalcont{These authors contributed equally to this work.}

\author[1]{Abhishek Bera}\email{abhishek.bera@uni.lu}

\author[1]{Antoine Richard}\email{antoine.richard@uni.lu}

\author[2]{Kazuya Yoshida}\email{yoshida.astro@tohoku.ac.jp}

\author[1]{Miguel Angel Olivares-Mendez}\email{miguel.olivaresmendez@uni.lu}

\affil[1]{\orgdiv{SnT-SpaceR}, \orgname{University of Luxembourg}, \orgaddress{\street{rue Richard Coudenhove-Kalergi}, \city{Luxembourg}, \postcode{L-1359},\country{Luxembourg}}}

\affil[2]{\orgdiv{Space Robotic Lab}, \orgname{Tohoku university}, \orgaddress{\street{Aoba Ward}, \city{Sendai}, \postcode{980-8577}, \state{Miyagi}, \country{Japan}}}

\newacronym{dymu}{DyMu}{Dynamic-Multi-Layered Path Planning}
\newacronym{esa}{ESA}{European Space Agency}
\newacronym{esric}{ESRIC}{European Space Resources Innovation Centre}
\newacronym{fps}{fps}{frames per second}
\newacronym{imu}{IMU}{Inertial Measurement Unit}
\newacronym{isru}{ISRU}{In-Situ Resources Utilisation}
\newacronym{lro}{LRO}{Lunar Reconnaissance Orbiter}
\newacronym{mrs}{MRS}{Multi-Robot Systems}
\newacronym{realms}{REALMS}{Resilient Exploration And Lunar Mapping System}
\newacronym{rgbd}{RGB-D}{RGB-Depth}
\newacronym{snt}{SnT}{Centre for Security, Reliability and Trust}
\newacronym{roi}{ROI}{region of interest}
\newacronym{rtabmap}{RTAB-Map}{Real-Time Appearance Based Mapping}
\newacronym{viper}{VIPER}{Volatiles Investigating Polar Exploration Rover}
\newacronym{vslam}{vSLAM}{Visual Simultaneous Localisation And Mapping}
\newacronym{v&v}{V\&V}{Verification and Validation}
\newacronym{ros}{ROS}{Robot Operating System}
\newacronym{dds}{DDS}{Data Distribution Service}
\newacronym{rmw}{RMW}{ROS MiddleWare}
\newacronym{hwmp}{HWMP}{Hybrid Wireless Mesh Protocol}
\newacronym{osi}{OSI}{Open Systems Interconnection}
\newacronym{kpi}{KPI}{Key performance indicator}

\abstract{Recent advancements in \gls{mrs} and mesh network technologies pave the way for innovative approaches to explore extreme environments. The Artemis Accords, a series of international agreements, have further catalyzed this progress by fostering cooperation in space exploration, emphasizing the use of cutting-edge technologies. In parallel, the widespread adoption of the Robot Operating System 2 (ROS 2) by companies across various sectors underscores its robustness and versatility. 
This paper evaluates the performances of available ROS 2 MiddleWare (RMW), such as FastRTPS, CycloneDDS and Zenoh, over a mesh network with a dynamic topology. The final choice of RMW is determined by the one that would fit the most the scenario: an exploration of the extreme extra-terrestrial environment using a \gls{mrs}. The conducted study in a real environment highlights Zenoh as a potential solution for future applications, showing a reduced delay, reachability, and CPU usage while being competitive on data overhead and RAM usage over a dynamic mesh topology.}



\maketitle

\section{Introduction}

\textcolor{red}{\acrfull*{mrs} have become popular} in recent years due to their potential for improving efficiency, flexibility, and robustness in various applications.
The application of \gls*{mrs} is already extensively studied in many fields such as agriculture~\cite{pretto_building_2021}, search and rescue~\cite{yan_survey_2013}, hospital logistics \cite{roldan_gomez_multi-robot_2019} or environmental monitoring \cite{espina_multi-robot_2011}.
However, research in \gls*{mrs} still requires extensive work and normalization of the existing technologies in the goal of reaching interoperability.

 \textcolor{red}{The Space industry focuses on \gls*{mrs}, especially with the growth of \gls*{isru}, which promotes \gls*{mrs} as a promising tool in a growing market\cite{baima_designing_2024}}. Future missions focus on exploring celestial bodies and require teams of robots to roam in extreme and unknown environments to provide a map and discover potential areas of interest. The work described in \cite{parker_multiple_2008} distinguishes four primary architectures for \gls*{mrs}: the Centralized approach, the Hierarchical approach, the Decentralized approach, and the Hybrid approach.
 Decentralized and Hybrid heterogeneous \gls*{mrs} are especially relevant for this context because such applications require redundancy, scalability, and adaptability. \textcolor{red}{However, those approaches face various challenges, such as more complex coordination due to their complexity.} 
 The work in \cite{yan_survey_2013} shows that a decentralized \gls*{mrs} offers way more scalability and adaptability than centralized \gls*{mrs}. By providing scalability and a better distribution of the resources, \gls*{mrs} offers a reliable approach for \gls*{isru} \cite{noauthor_overview_nodate}.

 \textcolor{red}{ Existing approaches even highlight the possibility of a \textit{coopetitive} system, where robots from various stakeholders are on the same mesh network \cite{chovet_trustful_2024}. These robots have the opportunity to collaborate or cooperate autonomously.}
\textcolor{red}{However, the space environment caters additional technical issues and constraints. Since the planetary surface is less cluttered with obstacles and interference, the coverage of a robot is larger. However, this implies communication delay and lower data overhead. 
Connectivity maintenance is the most crucial aspect for a space system, losing the connection to a robot is not acceptable since it couldn't be recovered \cite{staudinger_terrain-aware_2023}.}

Many companies are proposing robotic platforms that can perform several tasks, but interoperability is a need. 
The open robotic foundation proposes this through \gls*{ros} 2, the newest version of \gls*{ros}, focusing on aspects such as \gls*{mrs} and scalability \cite{macenski_robot_2022}. \gls*{ros} 2 is characterized by enhanced security features, real-time capabilities, and a flexible and dynamic architecture, allowing robots to interact in complex and dynamic environments. \textcolor{red}{Most of the features are brought by the use of the \gls*{dds} technology, acting as a \gls*{rmw} between ROS 2 and the network layer.}

\textcolor{red}{\gls*{dds} is a communication protocol often used in domains such as automotive, defence, finance, and simulation \cite{koksal_obstacles_2017}}. DDS plays the role of messenger that allows different parts of a system, such as robots in a \gls*{mrs}, to talk to each other effectively through mechanisms of subscription and publishment. \gls*{dds} can handle large amounts of data, deal with complex communication patterns, and is designed to work well even in challenging environments. Therefore, \gls*{dds} has become pivotal in \gls*{mrs} because it ensures that all the actors of a system are connected, updated, and working together seamlessly \cite{diluoffo_robot_2018}. 
However, alternatives to DDS are emerging, such as Zenoh \textcolor{red}{\cite{corsaro_zenoh_2023}} and propose newer approaches with announced better performances \cite{liang_performance_2023}.

Along with the \gls*{osi} model, ROS 2 and the middleware rely on the network layer to work. If many network architectures exist, exploration requires some specificity, such as scalability and the ability to react to topology changes. 
In the traditional star topology of wireless networks, each device connects to a central node. The mesh networks allow each device, or ``node", to communicate with all other participating nodes in the network either directly or via multi-hops. Therefore, it enhances the robustness of the network by offering multiple pathways for data transmission. Consequently, mesh networks are more reliable, as the failure of a single node rarely leads to a breakdown of the entire network. In addition, mesh networks are highly scalable, allowing the integration of more nodes without significant degradation in network performance \cite{sichitiu_wireless_2023}. This feature is critical in \gls*{mrs}, where the number of robots may vary depending on the task at hand.  \textcolor{red}{Despite these advantages, one of the biggest issues of the mesh networks lies in resource utilization. It requires more antennas to operate simultaneously compared to a centralized system where there would be one main antenna, leading to more power consumption. It also induces additional delay and potential consistency and synchronization issues.} \textcolor{red}{Middlewares such as DDS play a crucial role in \gls*{mrs}. To operate in extreme environments, these MRS need to be supported by meshed networks. Therefore, it is important to study and assess the impact of different middleware on such configurations. Hence, this study can guide the selection of the right \gls*{rmw} for a specific mission.}

This paper investigates the impact of using mesh networks and ROS 2 alliance for \gls*{mrs}-driven missions in extreme environments. The applied research focuses on a data-sharing scenario in extreme environments, implying some specific constraints such as long-distance communication, potential obstacles and topology changes. It also poses issues with energy consumption and sensing due to the unexpected environments.
The contributions of this work are as described below :

\begin{itemize}
    \item We compare the different \textcolor{red}{\gls*{rmw} technologies, presenting a Non-Line-of-Sight (NLOS) topology.}
    \item We present guidelines to choose the most suited \textcolor{red}{\gls*{rmw} depending the mission, for a \gls*{mrs} operating on a mesh network.} 
\end{itemize}

The paper is structured as follows.
The section \ref{sec:SOA} explores the state-of-the-art of \gls*{mrs} and Mesh Networks.
The section \ref{sec:research gap} highlights the needs of the state of the art.
The section \ref{sec:archi} details the robot agnostic architecture proposed.
The section \ref{sec:scenario} presents the scenario where this architecture would be applied.
The section \ref{sec:experiments} explains the experimental setup used for the quantitative study.
The section \ref{sec:results} displays the results.
Finally, section \ref{sec:conclusion} discusses the results and future works.

\section{State of the Art} \label{sec:SOA}

This section presents the details and state of the art of the various technologies involved in this paper.
\subsection{Decentralized MRS and ROS 2}

As explained in \cite{parker_multiple_2008}, we can distinguish four primary architectures of \gls*{mrs} :
\begin{itemize}
    \item Centralized: This approach coordinates all the robots from one single point. The solution focuses on the most optimized approaches since all the data for decision-making are known in one place. It also allows the use of simpler robots designed for highly specific tasks.
    \item Hierarchical: In this approach, an overseer is always overseeing a small group of robots, where some of them can also oversee a team. This allows the architecture to be strengthened against global failure.
    \item Decentralized: The decentralized approach allows each robot to make its own decisions independently. It is, by design, the strongest architecture but often leads to sub-optimal results and a higher difficulty in sharing complex tasks.
    \item Hybrid: This approach combines the previous approaches with the goal of allying robustness with efficiency.
\end{itemize}
Most of the studies done on the decentralized approach could also relate to the hybrid.
The recent improvements in service-oriented architectures (SOA) allow us to consider more economic interest in Sensing-as-a-Service and Robot-as-a-Service. \textcolor{red}{SOA is a design paradigm that allows services to communicate over a network through a loosely coupled relationship to facilitate business processes and data exchange}.  This would allow any client to use the functionalities of any robots available \cite{chen_robot_2010}. \textcolor{red}{ For example, a research institute on earth could use the service of a robot already present on-site instead of creating its own mission, creating economic interest for the research institute and the robot owners \cite{chovet_trustful_2024}}. This application facilitates special interest in extreme environments where missions are expensive and hardly reachable by humans, such as space exploration and \gls*{isru}.
However, due to the cost of launching and designing similar missions, it is expected to have multiple robots from multiple owners on astral bodies.
The robots from different owners still need a way to collaborate to propose an optimal SOA. However, the robots may remain from different companies and must keep their independence. This state has been studied since 1996 by \cite{brandenburger_co-opetition_2011} and is called coopetition. In the case of a Coopetiting system, the decentralized or hybrid architecture is the only architecture leaving enough freedom for each robot, or consortium to take its own decisions.



The authors in \cite{rizk_cooperative_2020} extensively studied cooperative heterogeneous and decentralized robotics and presented most of the current \textcolor{red}{applications}. They highlighted the main concerns of decentralized \gls*{mrs}, which are :
\begin{itemize}
    \item The Control architecture: How to ensure a global behaviour when robots only have local information. They classify three different controls: Reactive, Deliberative, and Hybrid.
    \item Consensus: Finding a consensus without a proper leader is still a research topic.
    \item Containment: The robot should be less interconnected to avoid a global failure.
    \item Formation: Robots must learn how to move in groups.
    \item Task Allocation: Most of the research focuses on this aspect.
    \item Intelligence: With the growing research in AI, the issue of how and if robots should share training data and learning is a hot topic.
    \item Optimisation: Due to the intrinsic suboptimal aspects of decentralized \gls*{mrs}, it remains one of the principal concerns. 
    \item Communication: Two main approaches exist: intrinsic, where robots communicate through sensors, and explicit, where the communication is done on a communication channel. Most applications rely on the second, while the other one is destined for specific use cases; it suffers from many issues, such as limited distance or the obligation to add sensors.
\end{itemize}

The authors insisted on the need for more research in the field of decentralized \gls*{mrs} \textbf{communication}. The main challenges in communication are:
\begin{enumerate}
    \item Efficiency
    \item Environment tolerance
    \item Consensus for interactions
    \item Optimizing Speed and Energy
\end{enumerate}

\textcolor{red}{In the case of space exploration, environment tolerance and optimization of energy consumption are the biggest challenges of decentralized \gls*{mrs}, where the centralized counterparts currently offer better solutions and more developed approaches.}


One of the first applications of a decentralized \gls*{mrs} was presented in \cite{asama_design_1989} with ACTRESS. ACTRESS is based on \textit{Universal Modular ACTOR formalism} by \cite{hewitt_universal_1973}. The system describes multiple autonomous components called robotor, which can communicate in peer to peer to any other robotor. This architecture has been designed to allow any autonomous architecture to join the system. This architecture already presents the interest of a decentralized \gls*{mrs} using a networking technology allowing peer-to-peer communication with its neighbours.


The author in \cite{parker_alliance_1998} presented ALLIANCE, a software architecture focusing on fault tolerance, often referenced as a basis for more recent works. It focused on scenarios where a robot loses the capability to achieve its tasks, and the other robots recover the task.

\textcolor{red}{The number of effective space robotic missions remains limited, but some scenarios have been proposed to study solutions to face the limitations posed by lunar exploration.}
The project LUNARES \cite{cordes_lunares_2011} involved a heterogeneous team of robots for autonomous data sampling and represents the interest of the scientific community for \gls*{mrs} in the context of space exploration.
\textcolor{red}{In the project ARCHES \cite{schuster_arches_2020}, researchers from the DLR demonstrated various technologies and collaborative processes during a space analogue mission at Mount Etna. They performed sample collections through an heterogeneous multi-robot collaboration, demonstrating the efficiency of multi-robotic teams for space missions. }

\textcolor{red}{With the objective of motivating companies and laboratories to explore new solutions for space resource exploration, the European Space Agency (ESA) and the European Space Resources Innovation Centre (ESRIC) created the Space Resources Challenge \cite{esa_esric_src_2023a}. Among the various implementations, REALMS2 \cite{van_der_meer_realms_2023} offered a decentralized approach to tackling space-related challenges such as communication delays.}


\textcolor{red}{To address the evolving needs of the robotic industry, the Open Robotics Foundation introduced ROS 2. 
Unlike its predecessor, ROS 2 is designed to provide reliable communication over distributed systems. This makes it particularly well-suited for MRS operating in harsh or remote conditions. 
ROS 2 uses a middleware-oriented network approach for high-performance inter-node communication, ensuring data integrity and timeliness crucial for coordinating complex robot tasks in unpredictable scenarios. ROS would require a centralized master to handle the system, making it hard to conceive a decentralized system without using workarounds. Thanks to the DDS, the default middleware for ROS 2, most ROS 2 nodes can communicate with any other node in the same network. DDS is a middleware protocol and API standard from the Object Management Group (OMG) for data-centric connectivity. Its core system is based on the Data-Centric-Publish-Subscribe approach and focuses on the real-time environment and scalable systems\cite{noauthor_about_nodate}. More details about the various middleware implementations of ROS2 can be found in section \ref{sec:DDS}.}

\textcolor{red}{Multiple works exist on \gls*{mrs} and ROS 2}; however, the focus is mostly on the control architecture \cite{rabbah_real_2021}, leaving the concern of communication with a lack of research. Often, the communication is assumed to be peer-to-peer with no flaw, which is far from what is expected in the context of extreme environment exploration \textcolor{red}{when it is highlighted that communication resource optimization is one of the biggest disadvantages of the decentralized system compared to the centralized approaches.}


\subsection{Mesh Networks}
Mesh networks are quite common in large-scale infrastructures and are the default topology constructed by separate entities connecting to each other. In the scenario of different robots owned by companies that share interests but not property over the system, the default architecture constructed by connecting all those companies will result in a mesh network. According to \cite{kc_wireless_2016}, the resulting network topology is the best trade-off between security, neutrality and the following points: 

\begin{itemize}
    \item Due to their decentralized nature, mesh networks are more \textbf{robust and reliable}. If one node fails or becomes disconnected, the network can dynamically reroute traffic through alternative paths, ensuring reliable connectivity.

    \item  Mesh networks can easily \textbf{scale} up by adding additional nodes without causing significant disruptions or requiring extensive reconfiguration. Each new node increases the network's capacity and coverage area.

    \item Even more, with multiple paths available for data transmission, mesh networks provide built-in \textbf{redundancy}. This redundancy enhances fault tolerance and minimizes the impact of node or link failures, improving network performance and operational time.
    
    \item Mesh networks excel in providing wide and \textbf{extended coverage}, particularly in large or challenging environments where traditional network setups may encounter limitations. The ability of nodes to communicate directly with one another facilitates broader coverage without relying on a centralized infrastructure.
    
    \item Mesh networks can be easily deployed and reconfigured to suit various scenarios due to their inherent \textbf{flexibility and adaptability}. They are suitable for dynamic environments where network nodes may be mobile or where the network's structure needs to be adjusted to accommodate changing requirements.

\end{itemize}

\gls*{mrs} can operate in various and challenging environments that require higher flexibility than traditional-centred networks. The extended coverage and resilience of that network are crucial points that \gls*{mrs} greatly appreciates, as each robot is considered an antenna expanding the area used by the robots. \textcolor{red}{Depending on the environment, rock formation can present obstacles to communication. However, with a fleet large enough,  the complex topology adaptability of the mesh can tackle these challenges. If a path becomes too long for the packages to go through, the mesh protocol can find a new path going through other robots}. Furthermore, this \textbf{flexibility and adaptability} of the network allow robots to be automatically reconnected to their pairs during a running experiment if the topology of the network changes. 

\textcolor{red}{As highlighted by the literature \cite{kc_wireless_2016,benyamina_wireless_2012}, one of the biggest issues with mesh networks is scalability since studies lack information about their behaviour for too large networks.}

In the following subsection, an introduction to the most commonly used mesh routing protocols will be achieved.

\subsubsection{AODV}
AODV stands for Ad-hoc On-demand Distance Vector, developed in 2003 as a low resource utilization routing protocol. It was designed with low network device resource usage and with the mindset to maximize packet communication between different devices \cite{chakeres_aodv_2004}.

\subsubsection{B.A.T.M.A.N Advanced}
B.A.T.M.A.N Advanced stand for Better Approach To Mobile Ad-hoc Networking (Advanced) and is a low-level decentralized routing protocol operating on the second \gls*{osi} layer. This routing protocol is a part of the Linux Kernel and only needs to be activated as a module to run on every device, even low-end ones \cite{jafri_split_2023}.
\\
Each Linux device, with the B.A.T.M.A.N Advanced module activated, will act as a node equipped with a switch over the network. This solution assumes that the devices used are running on an OpenWRT system, which provides a configurable and writable file system with package management.

B.A.T.M.A.N development was derived from the Optimized Link State Routing Protocol (OLSR), which is mostly like AODV proactive routing protocol \cite{jafri_split_2023}. The main difference between those two protocols is the solution used to pass information: Multipoint Relay (MP) in the case of OLSR and AODV and through available neighbours in the case of B.A.T.M.A.N.

\subsubsection{HWMP and HWMP+}

\textcolor{red}{\gls*{hwmp}, as described in \cite{noauthor_ieee_2007}, is defined by the IEEE standard 802.11s as a standard meshing protocol, which also defines the default metric as the airtime link.}
On each link, the metric $C_a$ described in \cite{yang_hwmp_2012} is computed as follow:
\begin{equation}
C_a = [O+\frac{B_t}{r}]\frac{1}{1-PER}
\end{equation}

With PER (Packet Error Rate):

\begin{equation}
PER = \frac{FrameResent}{FrameSent}
\end{equation}
Where $O$ represent the channel access overhead, $B_t$ is a frame of standard size, and $r$ is the transmission rate.
\newline
\newline
Later, \gls*{hwmp}+ was developed with the improvement of the PER computation in mind to get a more accurate estimation of the link quality and, as a result, allocate resources efficiently \cite{yang_hwmp_2012}. The authors of this solution \cite{yang_hwmp_2012} emphasized that \gls*{hwmp}+ is significantly more efficient than the previous solution, with an increase of 3\%-6\% additional receiving packets; hence, it generated 2.8\% to 10.4\% less routing overhead in general scenarios.
\
\\
The author of \cite{yang_hwmp_2012} highlighted that most of the metrics used for the Mesh system are less adapted to variable wireless links that can be affected by noises, interferences and even the limited energy of the nodes. The \gls*{hwmp}+ protocol proposed in this paper is designed with respect to a dynamic environment and real-time constraints.
\newline
\newline

Finally, Mesh Networks have already been experimented with in extreme environments. The winning team of the Darpa Subterranean challenge, Cerberus, used various robots dropping "breadcrumbs". \textcolor{red}{The "breadcrumbs" are relay nodes, made of an antenna and a battery, dropped in strategical places to ensure connection, creating an extended Mesh Network \cite{tranzatto_cerberus_2022}.}


\subsection{Exploring the performances of ROS 2}

The authors of \cite{maruyama_exploring_2016} studied the network properties of ROS 2 and its behaviour depending on various \gls*{dds} implementations. They point out that using the DDS technology makes it suitable for real-time distributed embedded systems and scalable. However, it also comes with an overhead since all ROS 2 messages must be converted to the DDS format before. This overhead and the resulting latencies are the main metrics studied in this paper.

It is important to note that this paper was published in 2016 and that the experiment was realized on an alpha version of ROS 2. Also, it focused on three DDS: FastDDS, OpenSplice, and Connext, while the current implementation of ROS 2 presents FastDDS, Connext, Cyclone, and Gurum. \textcolor{red}{And that Zenoh, a \gls*{rmw} different than DDS is foreseen to be implemeted. These RMW's are detailed in section \ref{sec:DDS}.} 

During various experiments, the authors of \cite{maruyama_exploring_2016} highlighted that ROS implied a loss of the initial message, whereas ROS 2 fixed this issue. 
They also presented different experiments to compare the three DDS. Using a DDS in ROS implies delay, which grows exponentially as the message size increases. The experiments showed that FastDDS could not share a message bigger than 256 bytes. However, it is also shown later that FastDDS uses fewer threads to run a ROS 2 node, the same amount as ROS1.

They concluded that even if ROS 2 implies an overhead compared to ROS, it can be overcome using its QoS Policy and not by using a master-node policy. They also highlight that FastDDS is, at that time, the most suited DDS application for real-time distributed embedded systems in terms of threads and memory.

\section{Research Gap} \label{sec:research gap}

As highlighted in state of the art, \gls*{mrs} is a widely studied topic, and ROS 2 is proposing a strong solution for developing and studying strong control architectures for various situations. However, there is often a lack of communication architecture, and the existing works are based on a hypothetical perfect case.
However, bigger and more complex \gls*{mrs} architectures can lead to bottlenecks and unexpected situations on unadapted networks.
More studies are needed on network architecture for robotic experiments and to propose some simple-to-implement architecture for any \gls*{mrs} application using ROS 2.

\section{Architecture Proposed} \label{sec:archi}

\subsection{Network Architecture}

\begin{figure}[htbp]
\centerline{\includegraphics[width=0.8\linewidth]{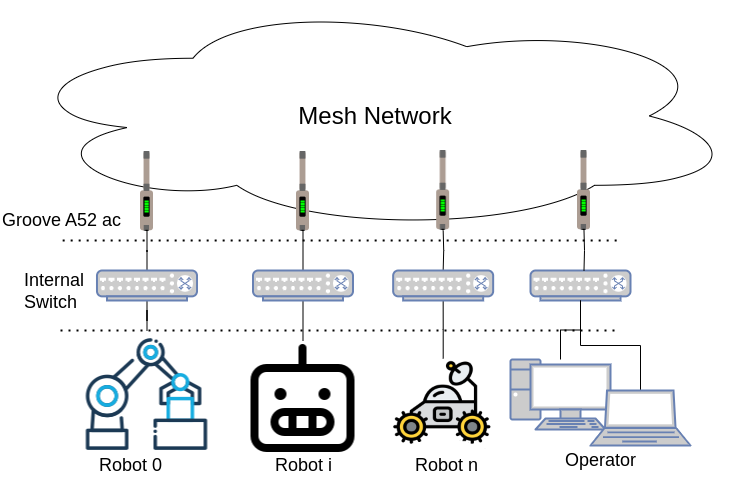}}
\caption{Mesh Network architecture}
\label{fig:mesh_archi}
\end{figure}

We consider a heterogeneous \gls*{mrs} deployed in an outdoor area to perform a set of tasks. A set $R = \{r_1,r_2,r_3,...,r_n\}$ consists of $n$ different robots connected through a mesh network $M$ and all participating in it. For $i \in \{1,n\}$, a robot $r_i$ is composed of $r_i = \{c_i,s_i,a_i\}$ with $c_i$ a set of $m$ embedded computers $c_i = \{c_{i_1},...,c_{i_m}\}$, a network switch $s_i$, and a mesh antenna $a_i$. \textcolor{red}{The embedded computers represent each device connected to the mesh that is directly on a robot}. The focus of this system is to propose high adaptability to all robotic platforms by connecting all the embedded computers to the switch, which is itself connected to the mesh router.
\textcolor{red}{We also consider a set of $l$ operators and each operator is represented as $o_k = \{c_k, s_k, a_k\}$, where $k\in \{1,l\}$, $c_k$ denotes associated computer, $s_k$ refers the associated network switch, and $a_k$ the respective antenna.} 
In that case, the mesh network comprises $n$ robots and one operator $o_k$  is represented by set $M = \{a_1, a_2, ..., a_n, a_k\}$. Figure \ref{fig:mesh_archi} presents the proposed architecture.

We consider that all robots have various capabilities, such as movement, data collection, and data treatment. They also have to face real-time mechanics.
As explained in the state of the art, \gls*{hwmp}+ is designed with large dynamic systems in mind.
According to \cite{yang_hwmp_2012}, \gls*{hwmp}+ has better metric computation, which results in less packet loss, reduced delay, and higher throughput compared to the standard \gls*{hwmp}. This is particularly noticeable in dynamic environments.
Among the different mesh protocols, \gls*{hwmp}+ presents the fitting properties for the architecture proposed here, which will be the mesh network used.


\subsection{ROS 2 Architecture} \label{sec:DDS}
\color{red}
Every embedded computer $c_{i_j}$ for $j \in \{1,m\}$ operates on a specific set of ROS 2 nodes that run the chosen \*gls{rmw} implementation. In its latest release, ROS 2 provides native support to two free DDS implementations, which are listed below:

\begin{itemize}
    \item \textbf{FASTDDS}: Also called FASTRTPS, this DDS has been used with ROS 2 since its beginning, always proposed as the default version. Since the implementation described in \cite{maruyama_exploring_2016}, there have been significant changes in its implementation, such as using shared memory transport, which allows the transport of larger messages \cite{noauthor_eprosima_nodate}. The authors announce a delay below $20\mu s$ for packets up to 15kB. According to their studies, in ideal conditions, they also present a throughput in the order of magnitude of $10^4 MB/s$.
    \item \textbf{Cyclone DDS}: Eclipse Cyclone DDS is designed as a free and open-source middleware. It is focused on high throughput and low delay, even if no official data are provided.
        The developers of FASTDDS have conducted a comparison with Cyclone, as cited in \cite{noauthor_fast_nodate}. For intra-process communication, FastDDS presents significantly better delay results, especially for larger payloads. In the case of inter-process communication, FastDDS claims to have a delay of around 30\% lower than Cyclone. They also highlight the stability of FastDDS, represented by the low delay increase as the payload gets larger. In terms of throughput, both DDSs present the same results.
\end{itemize}

\color{black}

\hfill

Another possibility, commonly used by \gls*{mrs}, is to rely on another middleware than DDS. \textcolor{red}{Eclipse Zenoh is very promising in this way, being close to natively implemented in the latest version of ROS 2. It offers the first easy to use \gls{rmw} not relying on DDS.} This technology is designed to streamline the discovery process and improve efficiency. During the discovery phase, robots attempt to identify the topics of other robots, which can be a crucial network operation. Fortunately, Zenoh has been developed to significantly reduce the overhead required for this process, cutting it down from 97\% to 99.9\%. This technology delivers results similar to those of DDS while minimizing the required resources.\cite{noauthor_minimizing_nodate}
Discovery overhead is frequently cited as a major problem with DDS technology in wireless networks, as it can cause network downtime.

In the paper \cite{zettascale_technology_improving_2022}, the authors explained that Zenoh is less likely to cause network failure compared to typical ROS 2 DDS. It is specifically created to operate smoothly even in unstable network conditions. One other advantage of Zenoh is the existence of pico Zenoh, a version compatible with microcontrollers. The only DDS implementation offering microcontroller-compatible features is FastDDS; however, various robotic applications use microcontrollers for motion. In the case of the usage of Zenoh in the architecture, all the \gls*{ros} 2 nodes within the robot $r_i$ should share the same domain id $D_i$, but the domain id of each robot $r_i$ should be different, $D_i \neq D_j$, $\forall \{i,j\} \in \{1,n\}(i\neq j)$.
\textcolor{red}{We note that all the available studies and comparisons \cite{noauthor_eprosima_nodate, noauthor_fast_nodate} use Gigabit Ethernet connections and no quantitative data are given for Wi-Fi, mesh networking or any other wireless technology such as 5g/6g.}

\section{Problem Scenario } \label{sec:scenario}
In our experiment, we only evaluate the performance of RMW. However, depending on the scenario, the ideal RMW might differ. This is why we created a fictional yet realistic scenario that will allow us to select the ideal RMW. For this fictional scenario, we consider a group of robots exploring a harsh environment, specifically a lunar environment. They extend the group capabilities by working together in a decentralised manner while also maintaining their independence. Each robot continuously generates and transmits sets of data called `messages'. As an example, these messages could represent small chunks of a point cloud acquired by various sensors. Multiple robots are expected to share their data (i.e. point clouds) for merging or optimisation operations \cite{chen_non-contact_2023,fotheringham_combining_2021}. In order to map even more efficiently, a robot can obtain the point cloud data of another robot to prevent the need to map an area already visited, thus conserving energy.
The state-of-the-art provides many mapping, task allocation, and control solutions \textcolor{red}{\cite{gomez_multi-robot_2022}}. \textcolor{red}{However, most applications are based on simulations or highly controlled networks, and the networking aspects are often considered out of the scope. Especially given the recent developments with ROS 2}.
The state-of-the-art needs more propositions of \gls*{mrs}-proof network architecture. Currently, many works and projects rely on access points and utilize the default settings of \gls*{ros} 2. \textcolor{red}{As stated in \cite{noauthor_distributed_nodate}, evolving network topologies remain an open question}. As a result, there is a need for a study about the network aspects of \gls*{mrs} relying on a decentralized architecture. In this context, we use \gls*{ros} 2 to implement a decentralized architecture. The main concerns of this problem to be surveyed are:
\begin{itemize}
    \item Data throughput
    \item Stability of a robot on the network
    \item Power consumption
\end{itemize}

\section{Experiments} \label{sec:experiments}
\subsection{Experimental Setup}
\subsubsection{Hardware}
Most of the embedded computers come with an internal antenna. The straightforward approach would be to use the embedded antenna and run the routing protocol on the embedded computer. However, this solution faces multiple issues.
\begin{itemize}
    \item Limitations: Internal antennas offer less data throughput and a limited range due to power and space limitations. 
    \item Resource usage: Embedded computers are optimized for specific purposes and are usually used for computationally intensive tasks such as image processing. Introducing a network protocol layer would increase the amount of computation required and make the system more susceptible to internal crashes.
    \item Bottlenecks: Robotic systems frequently incorporate multiple computers within the platform, all connected via Ethernet. However, relying on one computer to manage all networking tasks could result in significant data throughput limitations and reduce efficiency.
\end{itemize}

On the other hand, using an external router presents multiple advantages. The first one lies in using a dedicated internal board designed for network computation, allowing a better process of the packages and optimized routing. An external power source can provide a more significant data throughput and broader coverage. One of its key features is the ability to quickly implement any networking protocol, which is crucial when experimenting. Notably, using an external antenna can increase power consumption, which should be taken into account as a potential concern.

We use the \textit{Groove A52 ac} router from the brand \textit{Mikrotik} \cite{noauthor_mikrotik_nodate} as the mesh node for every agent of the \gls*{mrs}. This router offers more than 200 meters of coverage and is compatible with the norms 802.11ac (Wi-Fi 5). The router's design also meet the need for outdoor robotics, with a weatherproof design that easily fits all current robots. Finally, it can be powered over ethernet, providing a more straightforward wiring and power distribution approach.

The architecture proposed in Figure \ref{fig:mesh_archi} is highly scalable and adaptable to any robot. Since many robots rely on multiple embedded computers, they all need to be connected to a Switch connected to the Groove A52 AC. The router used in our setup is an \textit{EdgeRouter X} from the brand \textit{Ubiquity} set up as a switch \cite{noauthor_er-x-sfp_nodate}. However, any switch providing power over ethernet is suitable.

\begin{figure}[htbp]
\centerline{\includegraphics[width=0.5\linewidth]{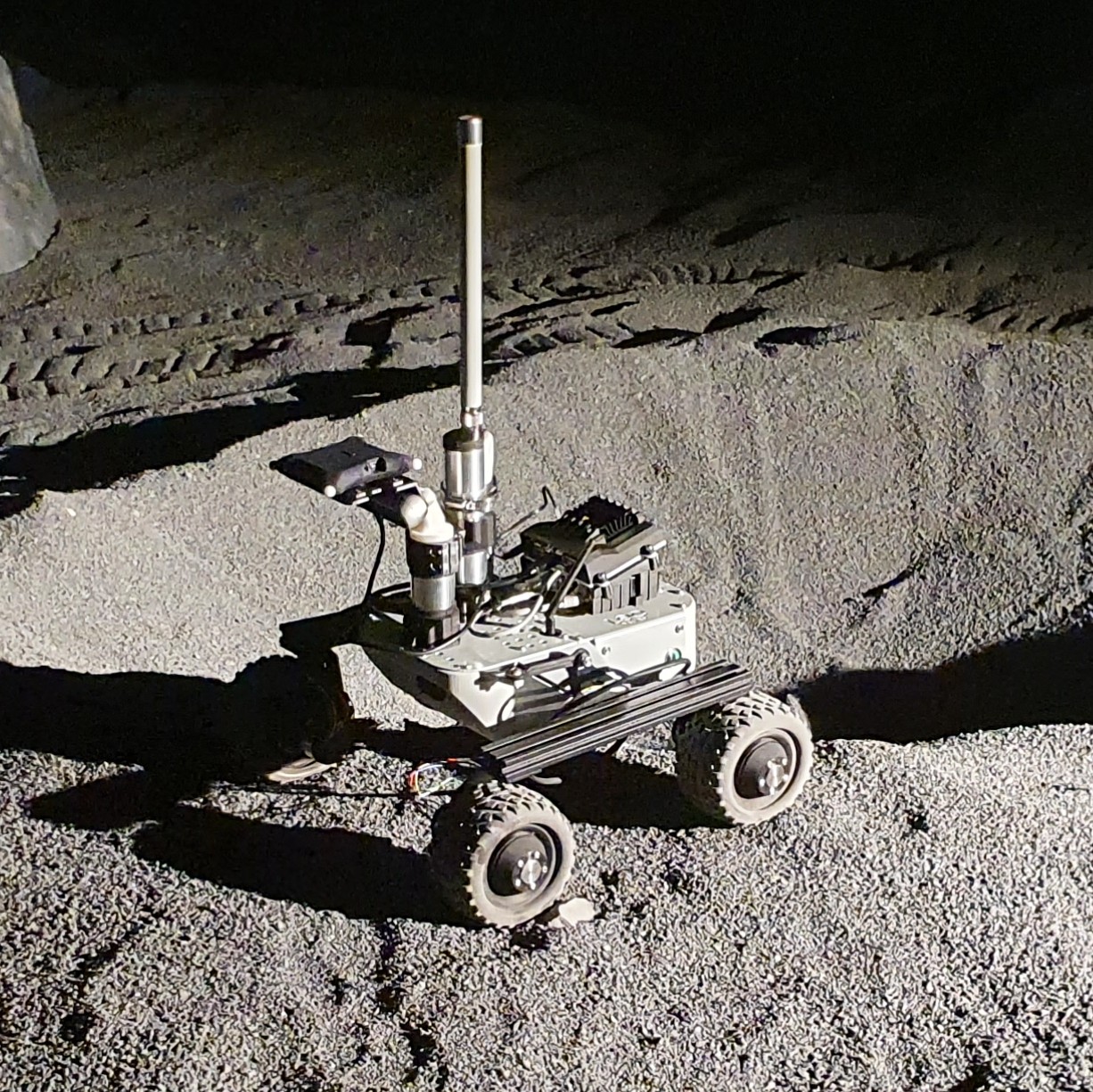}}
\caption{Experimental robotic platform}
\label{fig:leorover}
\end{figure}

The robots used for the experimentations are upgraded LeoRovers \cite{noauthor_leo_nodate} as in Figure \ref{fig:leorover}, equipped with an RGBD \textit{intel realsense\_d455} camera \cite{noauthor_introducing_nodate}, a \textit{RPlidar} \cite{peng_rplidar_nodate}, an embedded computer used for the computational heavy tasks \textit{Nvidia Jetson Xavier} \cite{noauthor_nvidia_nodate} and the setup previously mentioned.

The overall system is designated as Resilient Exploration And Lunar Mapping System 2 (REALMS2) and has been successfully tested in a real environment during the ESA-ESRIC Space resource challenge as detailed in \cite{van_der_meer_realms_2023}. This experiment benefit from this architecture however it would be possible to repeat the result only with a \textit{Raspberry Pi} and a \textit{Groove A52 ac} router.

\subsubsection{Software}
As explained in \cite{yang_hwmp_2012}, IEEE 802.11s defines \gls*{hwmp} as the default mesh protocol. The authors of \cite{yang_hwmp_2012} propose HWMP+ as a version more adapted to dynamic topology and optimising the shared channel resources. Most of the existing work was conducted in simulation and now requires further examination in mobile robotics.

\textcolor{red}{To establish the mesh network between the robots, Mikrotik provides instructions \cite{noauthor_hwmpplus_nodate}, creating a 2.4GHz Wi-Fi mesh network relying on the HWMP+ protocol. The \textbf{2.4GHz} band is chosen for its higher resistance of interference's and longer range. The authors of \cite{yang_hwmp_2012} propose HWMP+ as a version more adapted to dynamic topology and optimising the shared channel resources. Most of the existing work was conducted in simulation and now requires further investigation.}

All the network communications of ROS 2 are handled by a \gls*{rmw}, a middleware providing a publish-subscribe model for sending and receiving data, events, and commands among the nodes. The \gls*{rmw} solves many issues when it comes to distributing real-time data, sending and receiving data over a network, and providing consistency in the data model. It is acting on layers 5 and 4 of the \gls*{osi} network model. By definition, the mesh protocol assures the data link, so layer 2 of the \gls*{osi} model. This means that there is no difference in ROS 2 between a mesh network and a regular network.

All the embedded computers are running on Linux Ubuntu 22.04 and use ROS 2 Iron distribution.

\subsection{Metrics} \label{section_metrics}

\color{red}

We consider multiple metrics to compare and validate the more appropriate system.
The first set of metrics allows us to quantify the network the two crucial steps of the scenario, which is as follows:

\begin{itemize}
    \item \textbf{Data throughput}: Assessing the data throughput provided by the mesh network is crucial to fully understanding how different middlewares work together.
    \item \textbf{Jitter}: The delay measurements vary along a measurement. This variation is called the jitter and indicates the stability of the network. In a system with low jitter, messages consistently take about the same time to travel from sender to receiver.
\end{itemize}

\hfill

The second set of metrics are measured during the whole experiment, measuring the performances of each \gls*{rmw} implementation :

\begin{itemize}
    \item \textbf{Reachability}: Since the experiment consist on sending data from a robot to a lander, this metric evaluate when the robot is reachable from the lander. A device is considered reachable from another device if a connection can be established. 
    We define the average reachability $R(t)$ for a RMW as follow: 
    With $t \in [0,120], \tau_i(t) \in \{0,1\}$ :
    \begin{equation} \label{eq_reachability}
        R(t) = \frac{1}{N}\sum_{i=1}^{N} \tau_i(t)
    \end{equation}
    with $\tau_i(t)$ the reachability for a given run $i$, $N$ the total number of run for the RMW.
    The average reachability $R(t)$ is a floating-point value between 0 and 1. Higher values indicate a more stable connection.
    
    \item \textbf{Data overhead}: During the experiment, the robots will share messages of a fixed size over the network. However, each technology adds its own headers and additional content to the message. This metric studies the size of this overhead depending on the size of the message to quantify the network impact of each middleware.
    \item \textbf{Delay}: The delay is a crucial metric for real-time systems, representing the time it takes for a message to travel between a publisher and a subscriber.
    \item \textbf{CPU Usage}: Most embedded computers have limited resources, so monitoring the usage of the resources by the various middleware implementations is helpful.
    \item \textbf{RAM Usage}: Along with the CPU Usage, the RAM usage remains crucial since the memory is shared between many of the runnning processes and should remain available.
\end{itemize}

\color{black}

\begin{figure}[]
    \centering
    \begin{subfigure}{0.49\columnwidth} 
        \centering
        \includegraphics[width=1\linewidth]{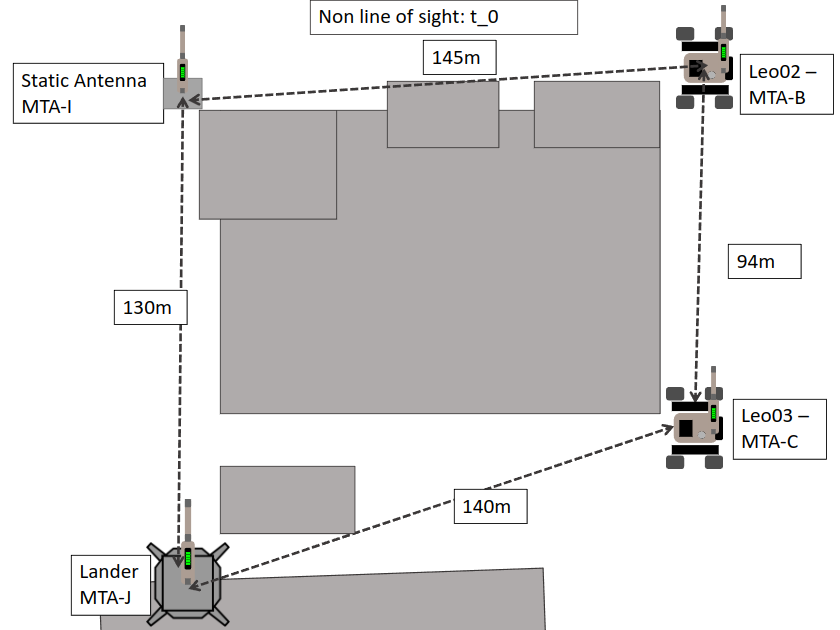}
        \caption{Initial configuration of the experiment.}
        \label{fig:nlos0}
    \end{subfigure}
    \hfill
    \begin{subfigure}{0.49\columnwidth}
        \centering
        \includegraphics[width=1\linewidth]{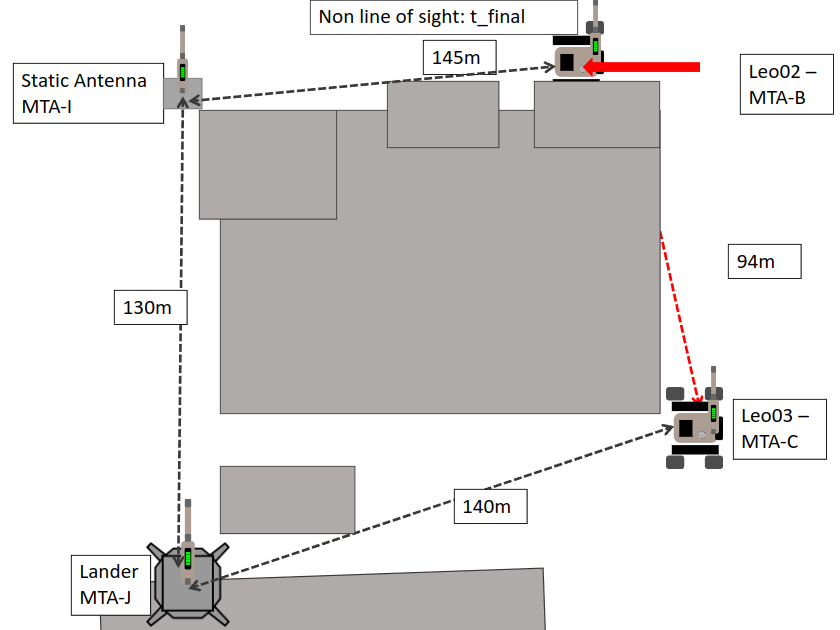}
        \caption{Final configuration of the experiment. No connection between Leo02 and Leo03}
        \label{fig:nlosfinal}
    \end{subfigure}
    \caption{Experimental Scenarios}
    \label{fig:exp1}
\end{figure}

\subsection{Experimental Scenario} \label{section_experimental_scenario}

\begin{figure}[ht]
    \centering
    \includegraphics[width=0.5\linewidth]{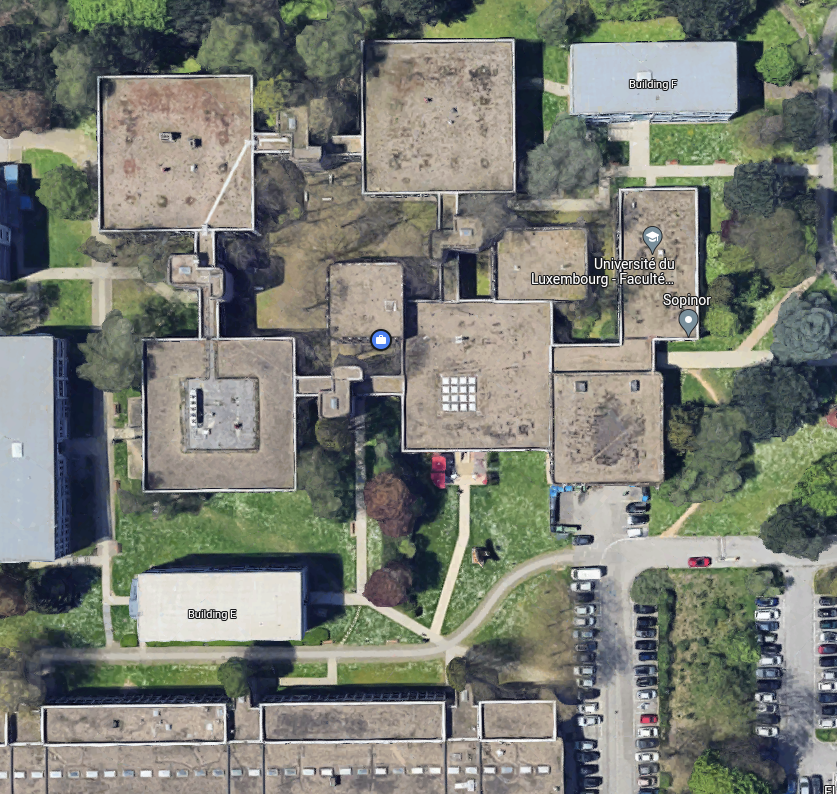}
    \caption{Urban terrain used for the first experimental scenario, the width of the visible environment is 175m}
    \label{fig:urban}
\end{figure}

As stated previously, ROS 2 handles communication through different middlewares. Since all the existing studies present results only in ideal conditions, we consider an approach that pushes the different middleware to their limits. As detailed in Figure \ref{fig:exp1}, the experiment involves two robots (Leo02 and Leo03), a lander, and a static antenna.\textcolor{red}{The chosen environment is signal-wise noisy and filled with obstacles in the urban environment (Fig. \ref{fig:urban}), pushing the network to its limits.}

\textcolor{red}{During the experiment, all used devices on the mesh network use IRON ROS 2 distribution. Furthermore, Leo02 is composed of two embedded devices, one internal and one external, as described in table \ref{table:hardware_rovers}. The internal onboard computer is used for low-level control (rover teleoperation) and is accessed out of the mesh network whereas the external computer is in charge of the communication of the data, and this is where the \gls*{rmw} changes take place. However, Leo03 is only used as a relay point.}

According to the authors in \cite{maruyama_exploring_2016}, quality-of-service (QoS) is one of the new technologies added by ROS 2 and significantly impacts network performance, especially with large data messages such as point clouds.
Any software using ROS 2 uses a QoS profile, and choosing the appropriate one can affect the networks. For this work, we consider the default QoS profile as only the RMW is evaluated and not the different available QoS.

\begin{table}[h!]
\centering
\begin{tabular}{lll}
\hline
       &  Internal         & External      \\ \hline
CPU    &  Raspberry Pi 4B  & Raspberry Pi 4B \\ 
Memory &  4GB              & 4GB            \\ 
OS     &  Ubuntu 22.04     & Ubuntu 22.04  \\
ROS Distro & Humble & Iron \\ \hline
\end{tabular}
\caption{Leo Rover composition}
\label{table:hardware_rovers}
\end{table}

\textcolor{red}{The experiment is divided into the three evaluated RMW that are each subdivided into seven fixed-sized messages (from a kilobyte to 64 kilobyte). This upper limit of 64 kilobytes has been found by preliminary tests, larger message size break the mesh network. To improve the quality of the result and get a proper sample, five successfully runs are completed per fixed-size messages leading to 3x7x5=105 runs. A run has a duration of 120 seconds and consists of Leo02 moving toward the static antenna while sending fixed-size messages to the Lander. This trajectory, fig.\ref{fig:nlosfinal}, has the purpose to change the network topology (because of the surrounding buildings) from a square shape (figure \ref{fig:topo_square}) to a line shape (figure \ref{fig:topo_line}) and investigate its impacts over each RMW. The messages sent on the network allow us to monitor the \gls*{kpi}s seen in section \ref{section_metrics}.}

\section{Results and Discussion} \label{sec:results}

\begin{figure}[!htbp]
\centerline{\includegraphics[width=1.2\textwidth]{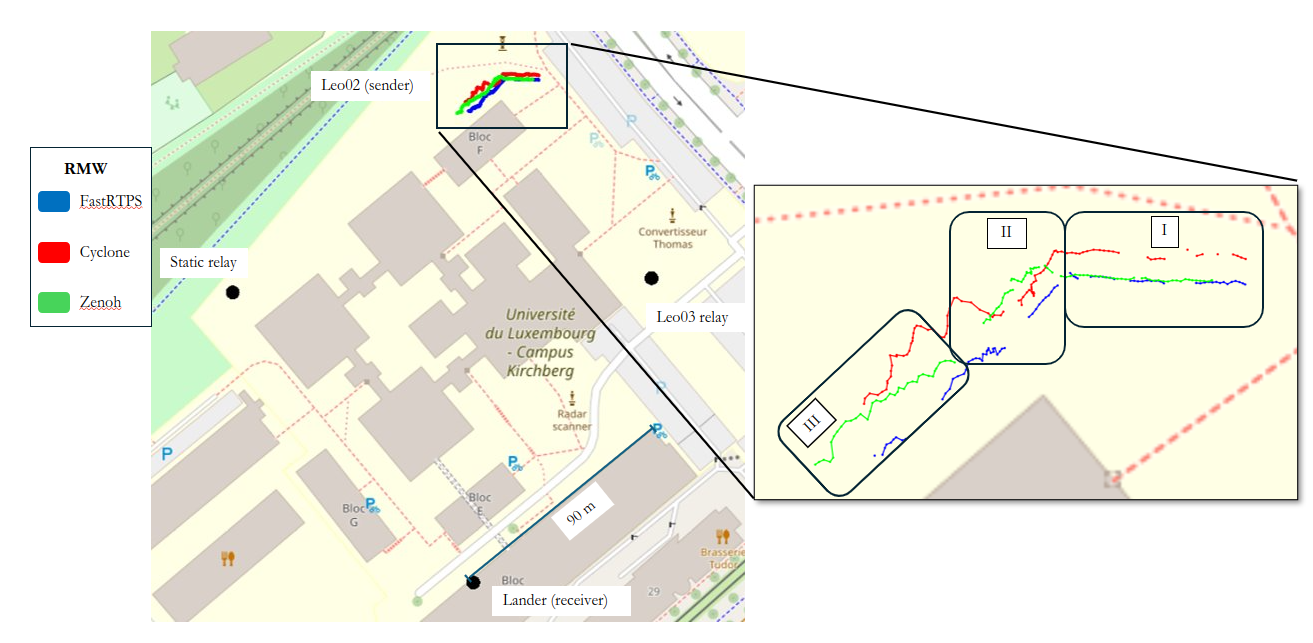}}
\caption{Global overview of the experiment}
\label{fig:global_overview}
\end{figure}

\begin{figure*}[!htbp]
    \centering
    \begin{subfigure}[b]{0.32\columnwidth}
        \includegraphics[width=\textwidth]{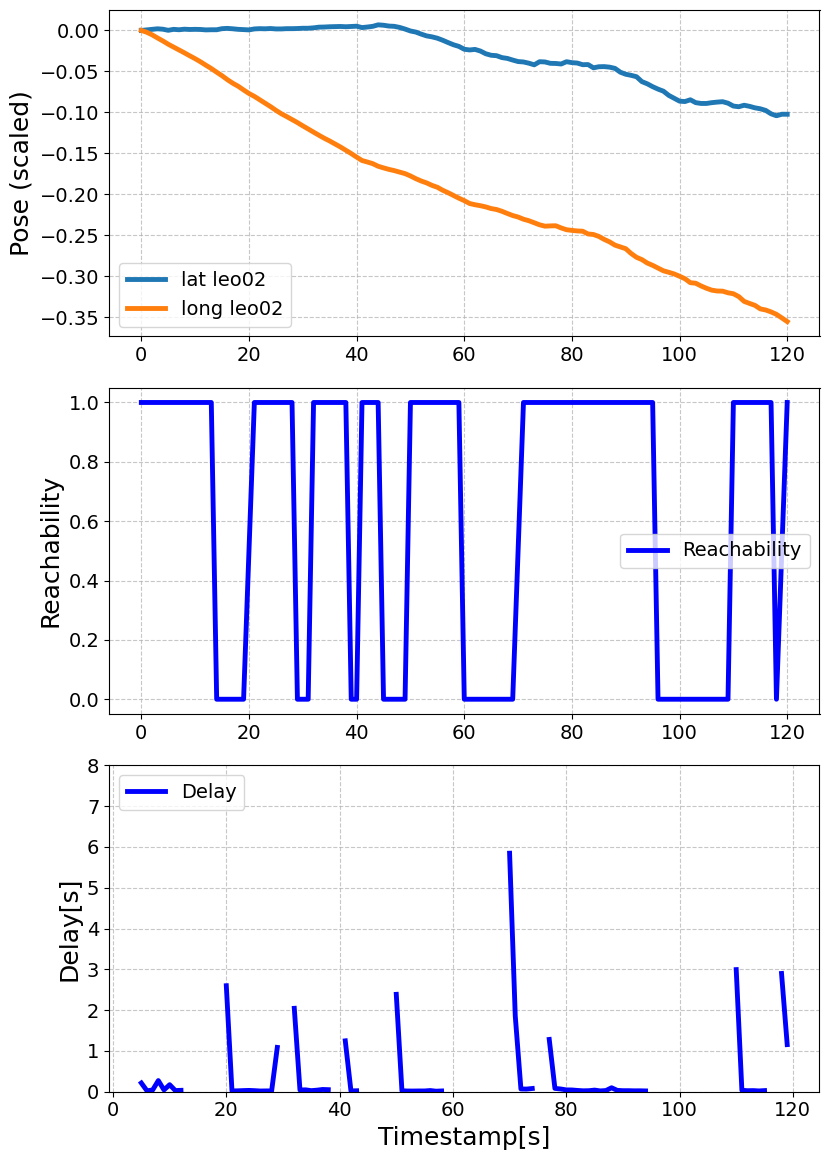}
        \label{fig:overview_fast}
        \caption{Fifth run of Fast DDS}
    \end{subfigure}
    \begin{subfigure}[b]{0.32\columnwidth}
        \includegraphics[width=\textwidth]{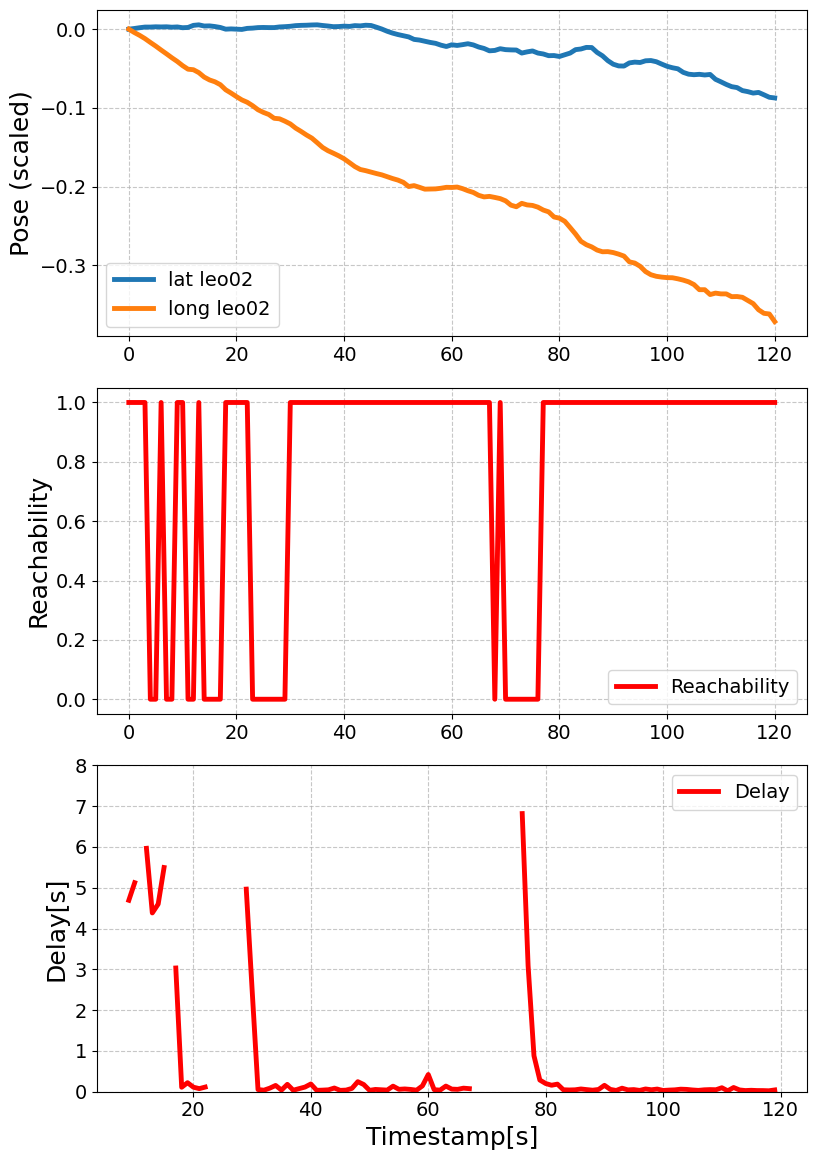}
        \label{fig:overview_cyclone}
        \caption{First run of Cyclone DDS}
    \end{subfigure}
    \begin{subfigure}[b]{0.32\columnwidth}
        \includegraphics[width=\textwidth]{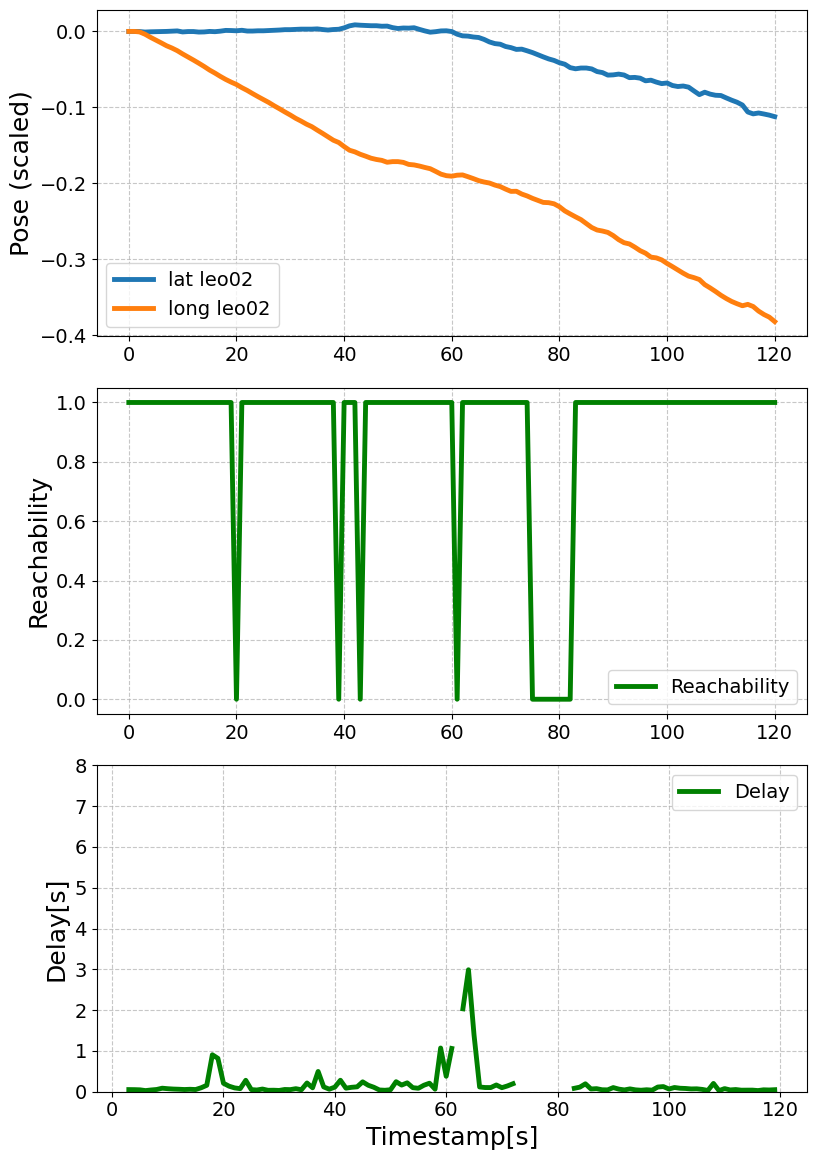}
        \label{fig:overview_zenoh}
        \caption{Fifth run of Zenoh RMW}
    \end{subfigure}
    \caption{Overviews of a singular run for each RMW for a message size of 16 KiloBytes. Trajectory, reachability and delay are showed}
    \label{fig:overview_network}
\end{figure*}

\textcolor{red}{As explained in section \ref{section_experimental_scenario}, each RMW has been studied with seven message types over five distinct runs. Figure \ref{fig:global_overview} demonstrates an overview of the experiment, including the robot's trajectory given each RMW. Figure \ref{fig:overview_network} presents the network overview of each of these singular runs. Both figures emphasise the connection lost between 60 and 90 seconds, allowing the network to change its topology.}

\begin{figure}[!ht]
    \centering
    \begin{subfigure}{0.49\columnwidth} 
        \centering
        \includegraphics[width=1\linewidth]{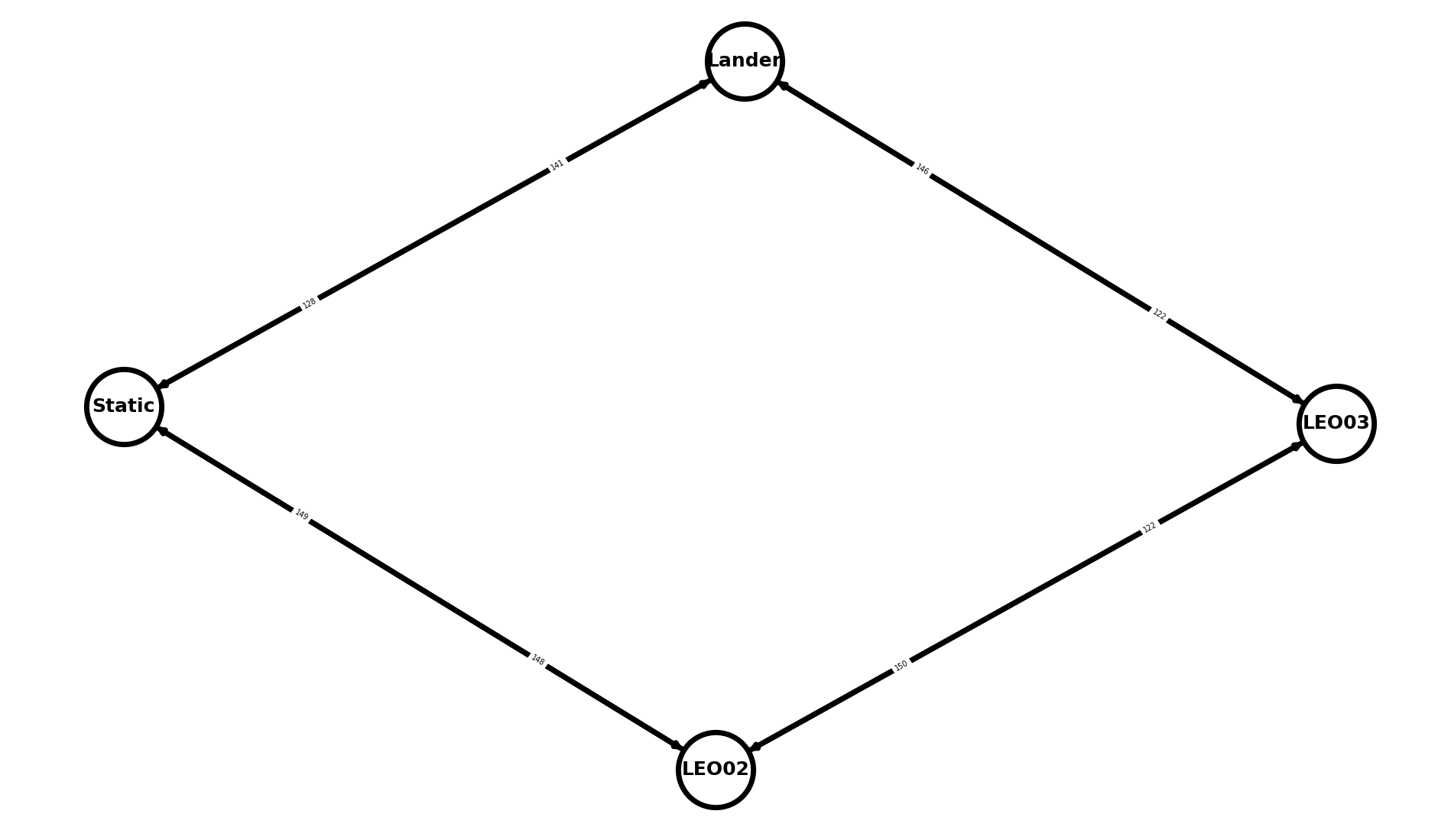}
        \caption{Square network topology (beginning of experiment)}
        \label{fig:topo_square}
    \end{subfigure}
    \hfill
    \begin{subfigure}{0.49\columnwidth}
        \centering
        \includegraphics[width=1\linewidth]{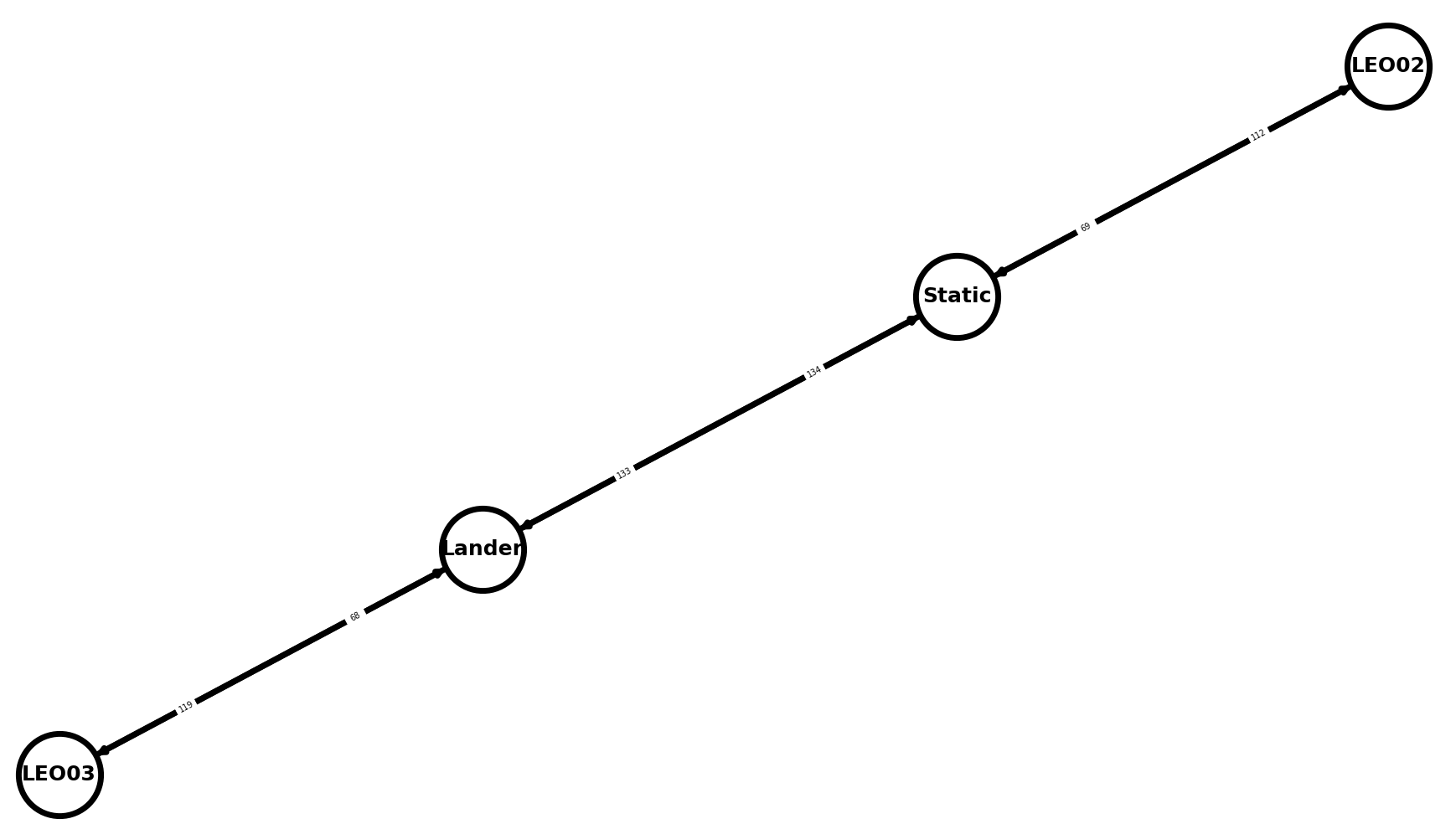}
        \caption{Line network topology (end of experiment)}
        \label{fig:topo_line}
    \end{subfigure}
    \caption{Result of the network topology analysis during the experiment done with a GUI. A circle represent a node of the mesh network (antenna) connected to its direct neighbour.}
    \label{fig:topo_exp1}
\end{figure}

\textcolor{red}{The figure \ref{fig:global_overview} has a zoom on the trajectories of each RMW. The missing values in the trajectories are representing the loss of connection of the rover. During those `blackout' moments, the network is either saturated or in reconfiguration. A closer look to the trajectories allow us to identify three different phases. During the first phase (I), the rover is connected to the mesh network in a full square topology (see figure \ref{fig:topo_square}) and therefore enhance good communication of the packages over the mesh allowing multiple routes. The second phase (II) is the transition phase where the mesh change it's topology due to the moving antenna on Leo02. During this phase, loss of connection, delayed messages and unstable network is expected. Finally, the third phase (III) shows a reconfiguration of the network in `line' (see figure \ref{fig:topo_line})}.

\textcolor{red}{In addition to the map overview, the figure \ref{fig:overview_network} shows how the network behave during the experiment. The top graph represents the position of the rover related to its starting point, the middle graph is a metric called `reachability' (if the rover is connected to the network or not), and lastly, the delay. Missing data in the delay graph are due to connection loss between the rover and the mesh network.}

 \textcolor{red}{A GUI tool helps us to keep an overview of the network topology during the experiment. Figure \ref{fig:topo_square} displays the topology at the beggining of the experiment (t=0), while figure \ref{fig:topo_line} displays the topology at the end of the experiment (t=120s).
This 120 seconds duration was decided after measuring the time for the network to change its topology while driving at a speed of 0.5m/s. After measurement, 75 seconds was enough to reorganized the network. However, all the run have a duration of 120 second to fully understand the behavior of the network before and after the change in topology.
Table \ref{table:network_metrics} presents the average capabilities of the mesh network in the beginning of the scenario (t=0) and at the end of the scenario (t=120).}

\begin{table}[!ht]
\centering
\small 
\setlength\tabcolsep{3pt} 
\begin{tabular}{|l||c|c|c|c|}
\hline
\small Scenario & \multicolumn{2}{c|}{\small t=0} & \multicolumn{2}{c|}{\small t=120} \\ \hline
                          & \scriptsize lander to leo02 & \scriptsize leo02 to lander & \scriptsize  lander to leo02 & \scriptsize leo02 to lander \\ \hline
\small Data throughput (TCP)                 &     270 Kbits/s            &  164 kbits/s               & \small 1.31 Mbits/s      &  \small 1.15 Mbits/s               \\ \hline
\small Jitter (UDP)                   &     0.000ms            &   245.385ms             &  0.000 ms              & 7.850ms               \\ \hline
\end{tabular}
\caption{Network Metrics over the mission without ROS 2}
\label{table:network_metrics}
\end{table}

Since the \textbf{packet loss} and \textbf{delay} of the ping command do not vary in any scenario, they are not included in the table.
\textcolor{red}{It is noticeable in Table \ref{table:network_metrics} that the distance between the robot and the relay impact the data throughput and that relays should be used carefully. This behaviour can be explained by the number of packet losses: the further, the antennas are spread, the higher the chances of packet loss. Therefore reduced usable bytes are received per second.}

In the scenario where node A and node B want to communicate, we learn that if node C wants to offer relaying services, it should only get concerned when A and B are on the edge of the direct communication span.
\textcolor{red}{It is also noticeable in table \ref{table:network_metrics} that the jitter is highly impacted by the distance between each node. Also, using the lander as an emitter provides better results than the robot. Those results can be explained by the lander's antenna being attached at the first floor of a building and therefore offering a better coverage.}

\textcolor{red}{One hundred five runs were initially expected; however, we encountered some failures during the experiment, leading us to perform 136 runs instead, giving a success rate of 77\%. It is noticed that this success rate varies depending on the RMW, 73\% for fast, 77\% for cyclone and 83\% for zenith.}

\textcolor{red}{The following subsections describe the evaluations of the studied \gls*{kpi}s: reachability, delay, data overhead, CPU usage, and RAM usage over time.}

\begin{figure}
    \centering
    \includegraphics[width=\textwidth]{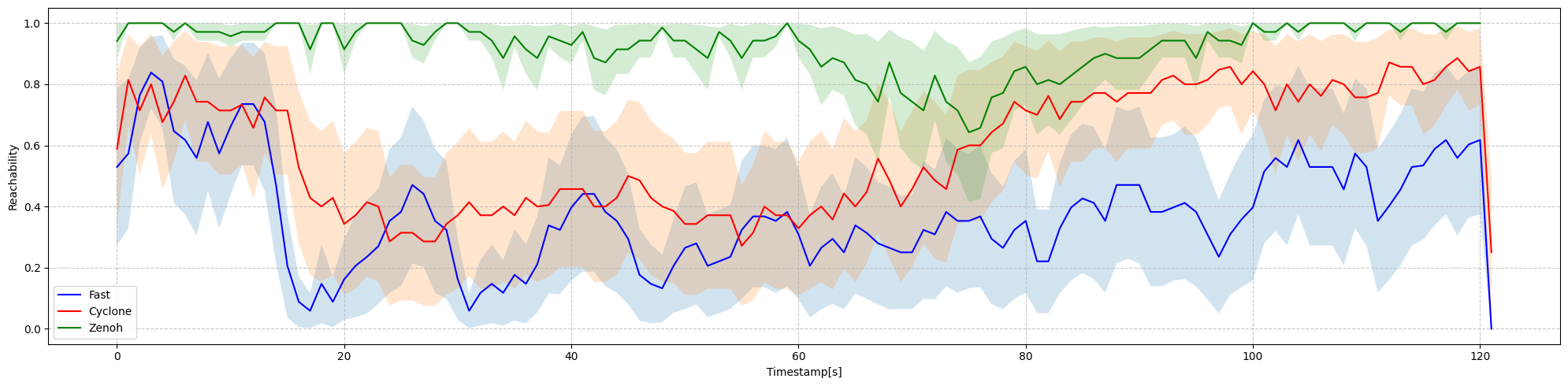}
    \caption{Average reachability over all the runs for each RMW with the variance.}
    \label{fig:avg_reach}
\end{figure}

\textcolor{red}{\subsection{Reachability}}

\begin{figure}[h]
    \centering
    \includegraphics[width=\textwidth]{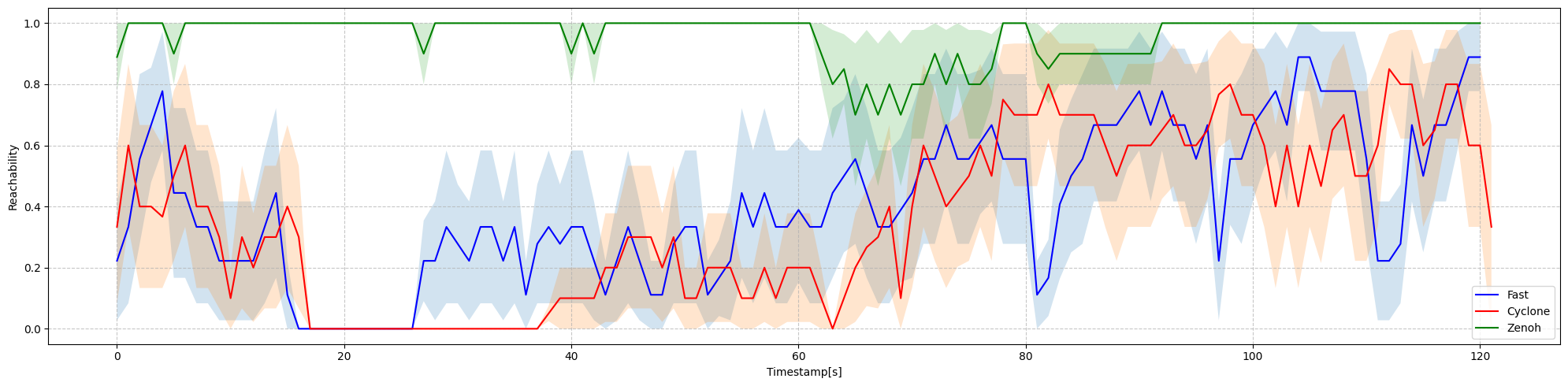}
    \caption{Average reachability for messages of sizes one and two kilobytes}
    \label{fig:avg_reach_1}
\end{figure}

\begin{figure}[h]
    \centering
    \includegraphics[width=\textwidth]{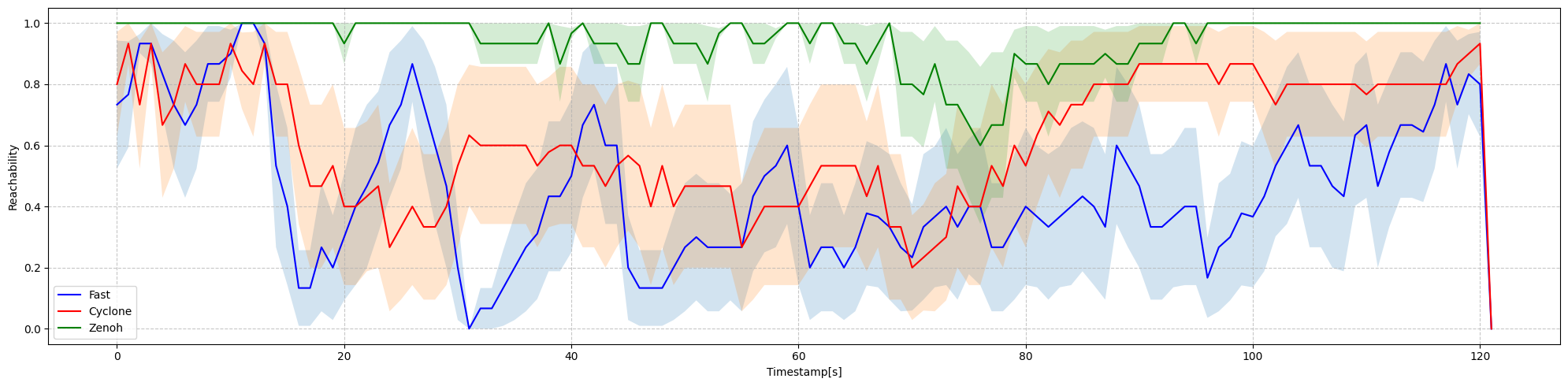}
    \caption{Average reachability for messages of sizes four, eight, and sixteen kilobytes}
    \label{fig:avg_reach_4}
\end{figure}

\begin{figure}[h]
    \centering
    \includegraphics[width=\textwidth]{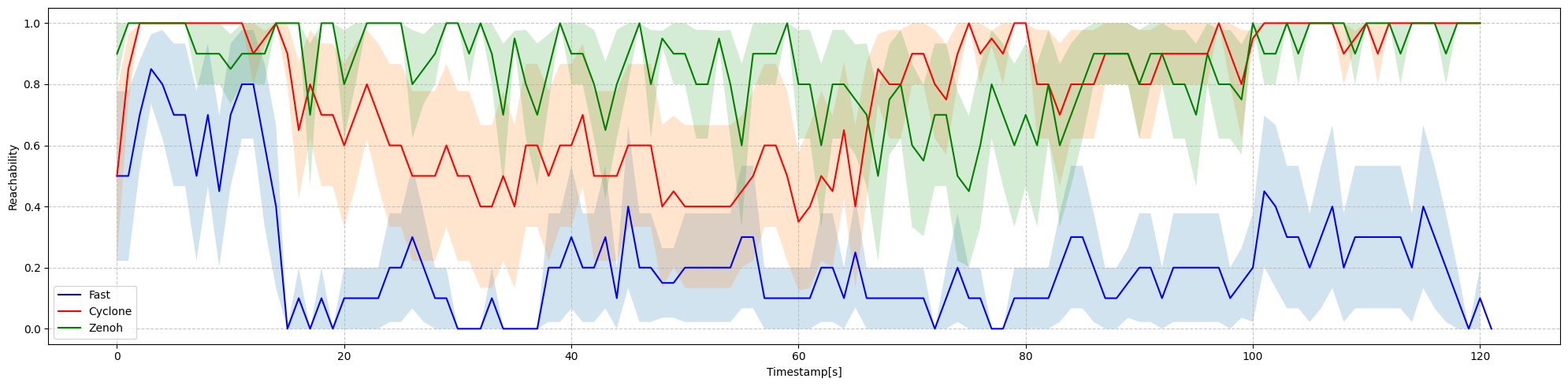}
    \caption{Average reachability over three packs of message size for each RMW with the variance.}
    \label{fig:avg_reach_32}
\end{figure}


In our scenario, the reachability of Leo02 from the lander is evaluated. 
Based on the introduced formula for the reachability, equation \ref{eq_reachability}, $N$ is the total amount of run per RMW (i.e 35), $\tau_i(t) \in \{0,1\}$ represent the reachability of Leo02 at time $t$ and $R(t)$ is the average reachability of a RMW at a time $t$. 
Figure \ref{fig:avg_reach} displays the average reachability $R(t)$ with its variance for each RMW given time.

\textcolor{red}{The data reveals a similar behaviour for Fast and Cyclone, all relying on the DDS technology; however, Zenoh is showing significant improvements in terms of network stability.}
\textcolor{red}{The Figures \ref{fig:avg_reach_1},\ref{fig:avg_reach_4},\ref{fig:avg_reach_32} shows the average reachability for three different message groups. Figure \ref{fig:avg_reach_1} corresponds to the smallest messages: one and two kilobytes. Figure \ref{fig:avg_reach_4} highlights the medium messages (4, 8, and 16Kb), while Figure \ref{fig:avg_reach_32} shows the bigger messages (32 and 64Kb). Both small and medium messages as shown in Figure \ref{fig:avg_reach_1} and \ref{fig:avg_reach_4} emphasize on Zenoh that outperform the DDS (Fast and Cyclone) in terms of reachability. However, it is interesting to say that the results for bigger messages shown in Figure \ref{fig:avg_reach_32} mitigate the performance of Zenoh with comparable results with Cyclone. On the other hand, Fast is underperforming compared to the other RMW.}
\\
\subsection{Delay}

\textcolor{red}{In this section, the delay between each message received is studied. For reference, a message is sent by Leo02 every 0.5 seconds on the network but might be lost or corrupted during the transfer. This is why the delay in receiving the messages is studied; a higher delay means more time to update a potential point cloud or outdated information about the respective rover's residual energy.}
\begin{figure}[h!]
    \centering
     \begin{subfigure}[b]{\columnwidth}
        \includegraphics[width=\textwidth]{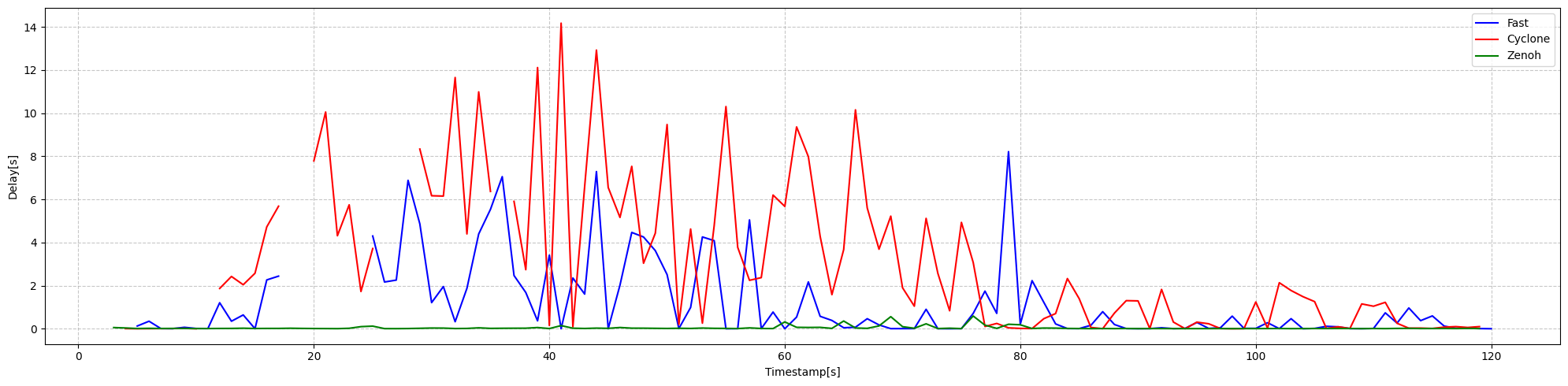}
        \caption{Average delay for messages of sizes one and two kilobytes}
        \label{fig:avg_delay_1}
    \end{subfigure} \\
    \begin{subfigure}[b]{\columnwidth}
        \includegraphics[width=\textwidth]{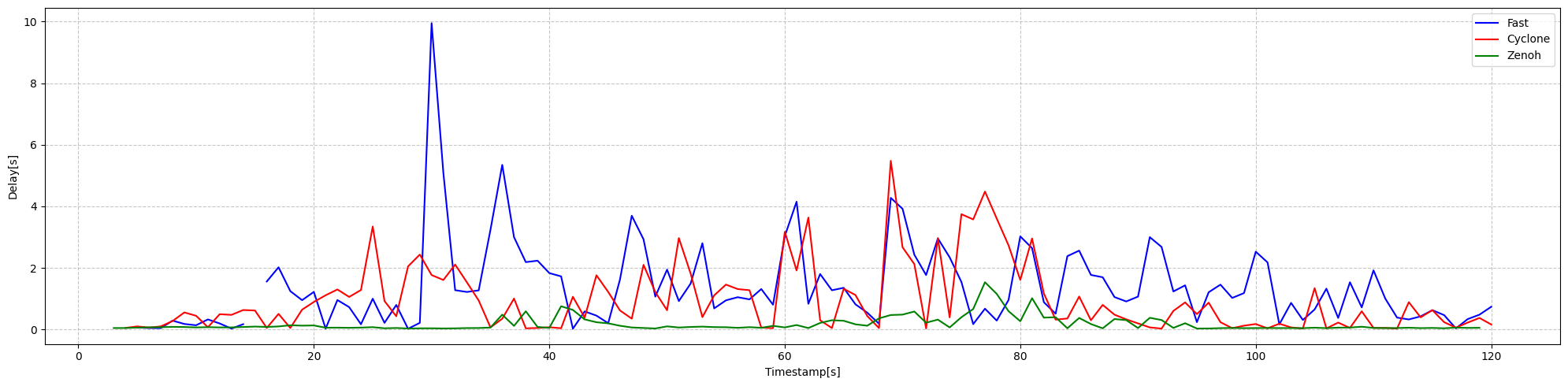}
        \caption{Average delay for messages of sizes four, eight, sixteen kilobytes}
        \label{fig:avg_delay_4}
    \end{subfigure} \\
    \begin{subfigure}[b]{\columnwidth}
        \includegraphics[width=\textwidth]{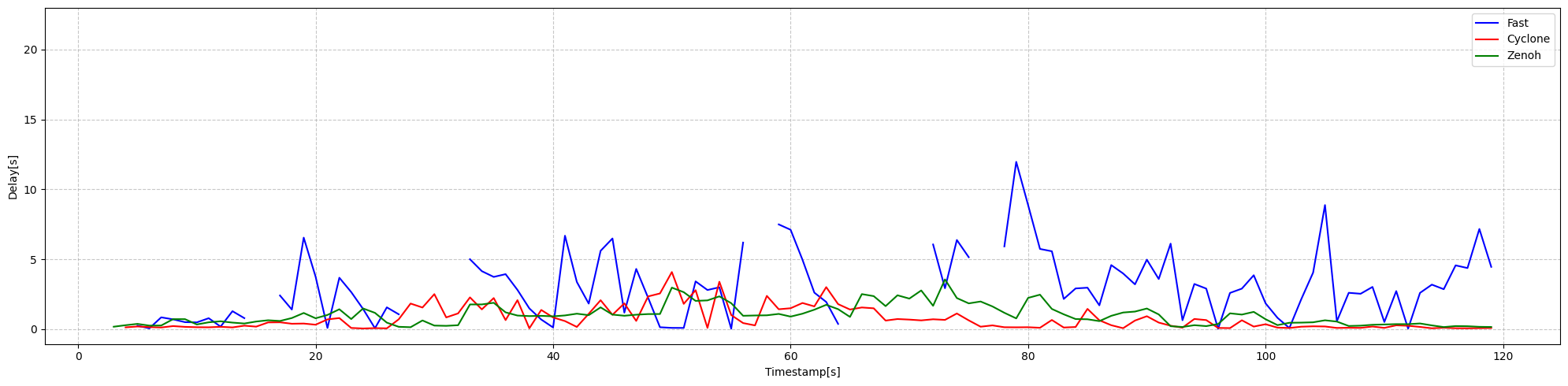}
        \caption{Average delay for messages of sizes thirty-two and sixty-four kilobytes}
        \label{fig:avg_delay_32}
    \end{subfigure} \\
    \caption{Average delay in second over three packs of message size for each RMW without the variance.}
    \label{fig:avg_delay}
\end{figure}

\textcolor{red}{Figure \ref{fig:avg_delay}, shows the average delay in seconds for various message sizes transmitted using the different middleware implementations. Compared to reachability, the variance is not displayed to enhance readability (extremums leads to flatten the graph). However, it is notable that FastDDS presents way more variance than Zenoh or Cyclone, reaching values up to a fifty-second delay. Figure \ref{fig:avg_delay_32} shows that Zenoh outperforms the DDSs by an order of scale. It is also interesting to notice that FastDDS performs better than Cyclone for very small messages, which follows FastDDS's claims. On the other hand, Zenoh has comparable results to Cyclone for large messages following the reachability's tendency.}
\textcolor{red}{It is also important to notice that Cyclone and Fast seem to be impacted by a less stable network while Zenoh remains stable all along. Around the end of the experiment, when the data throughput exceeds 1Mb/s (table.\ref{table:network_metrics}), both DDSs seem more stable.}

In the context of \textcolor{red}{extra-terrestrial} extreme environments, the delay of communication is a less crucial parameter since \textcolor{red}{any delay induced by the network would be negligible compared to the delay between the operator and the robot}.  For example, in the case of lunar exploration, we can expect a delay between the Earth and the moon of 2 seconds. \textcolor{red}{Such delay already removes the robot's real-time teleoperation capabilities.}

\textcolor{red}{\subsection{Data overhead}}

\begin{figure}[htbp]
    \centering
    \includegraphics[width=\textwidth]{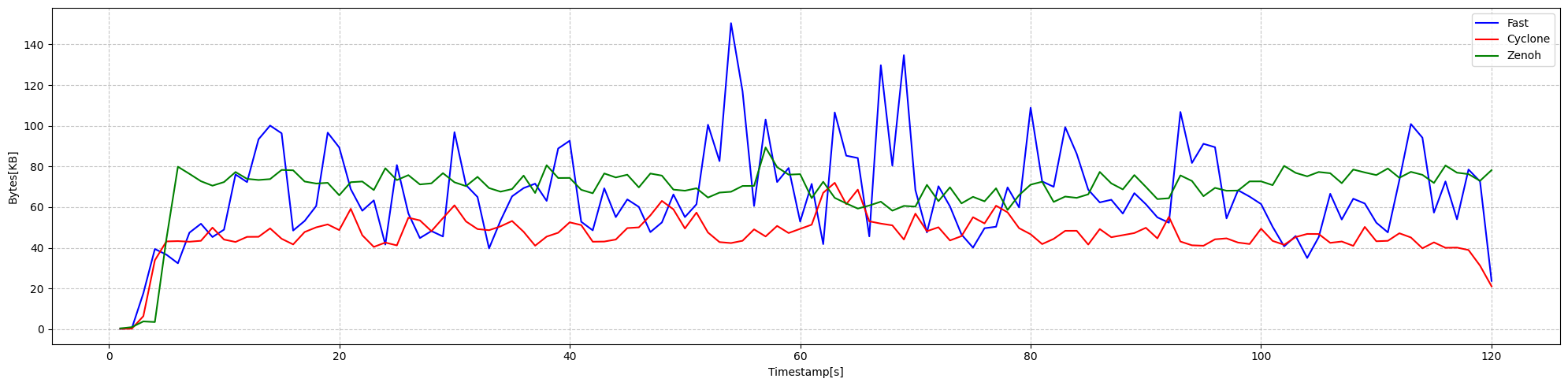}
    \caption{Average bytes sent by leo02 over all the runs for each RMW without the variance.}
    \label{fig:avg_bw}
\end{figure}

\begin{figure*}[htbp]
\begin{subfigure}[b]{0.49\textwidth}
    \includegraphics[width=\textwidth]{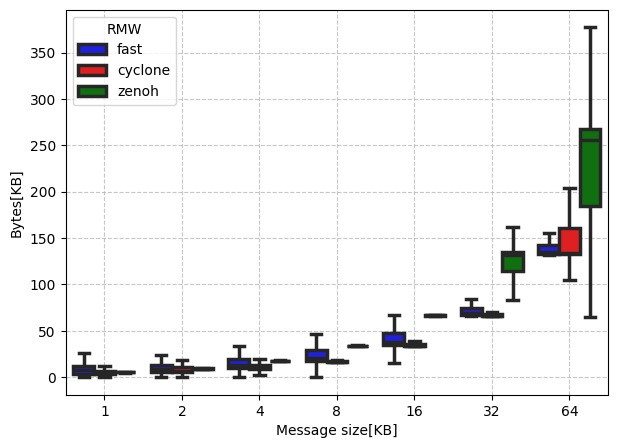}
    \caption{Data overhead of leo02's ROS 2 related processes for every fixed size message}
    \label{fig:leo02_bw}
\end{subfigure}
\hfill  
\begin{subfigure}[b]{0.49\textwidth}
    \includegraphics[width=\textwidth]{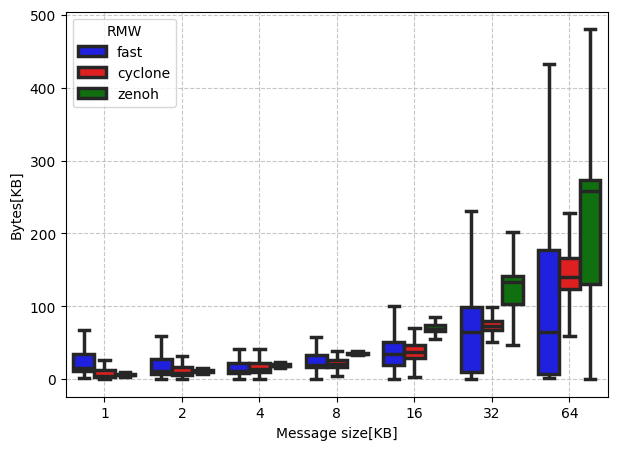}
    \caption{Data overhead of the lander ROS 2 related processes for every fixed size message}
    \label{fig:lander_bw}
\end{subfigure}
\caption{Box Plot representing the data overhead, for each RMW at every fixed size}
\label{fig:box_plot_bw}
\end{figure*}

\textcolor{red}{The data overhead is characterised by two metrics: the bytes sent by Leo02 and the bytes received by the lander. Since for all the RMWs, the messages are of the same size, measuring the total bytes sent gives an overview of the message overhead of each RMW. Figure \ref{fig:avg_bw} shows the average bytes sent by leo02 over the different RMW. As showed in the graph, this value does not fluctuate a lot except for Fast during a run. These peaks may be one of the reasons for the largest network instability led by Fast. Since the average bytes sent by each RMW do not fluctuate much over time, Figure \ref{fig:box_plot_bw} offers a Box Plot graph featuring the bytes sent by Leo02 for each message size and RMW \ref{fig:leo02_bw} along with a Box Plot featuring the bytes received by the lander for each message size and RMW }
\ref{fig:lander_bw}.  

\textcolor{red}{Figure \ref{fig:box_plot_bw} provides a better understanding of the impact of each RMW over different message sizes. The results of the lander and leo02 are comparable as expected, except for a slight increase of bytes on the lander side (scale differs on both graphs; Figure \ref{fig:lander_bw} has a larger scale). Zenoh has overall greater data overhead consumption in both figures for larger messages. However, Zenoh has a more limited range for small and medium message sizes, whereas the other RMW tends to fluctuate more.}

\textcolor{red}{Based on those results, Zenoh appears to be more predictable for small and medium messages, whereas Fast and Cyclone tend to lead to more inconsistency and, therefore, more network instabilities.}

\subsection{CPU and RAM usage}

\textcolor{red}{In this section, we discuss the CPU and RAM usage of ROS 2 processes. It is important to emphasize the running processes on the rover. We only consider the packages designed for the experiment to be running (sender for leo02 and receiver for lander), excluding even the ROS daemon. As an embedded system has a limited amount of resources, this metric allows us to quantify the impact of each RMW on the system. An RMW taking more resources leaves less room for other algorithms that might be running on the system, such as SLAM, navigation, or exploration. Furthermore, taking more resources has a direct impact on the power consumption of the embedded device and, therefore, the battery life of the robot.}

\begin{figure*}[htbp]
    \begin{subfigure}[b]{0.49\textwidth}
        \includegraphics[width=\textwidth]{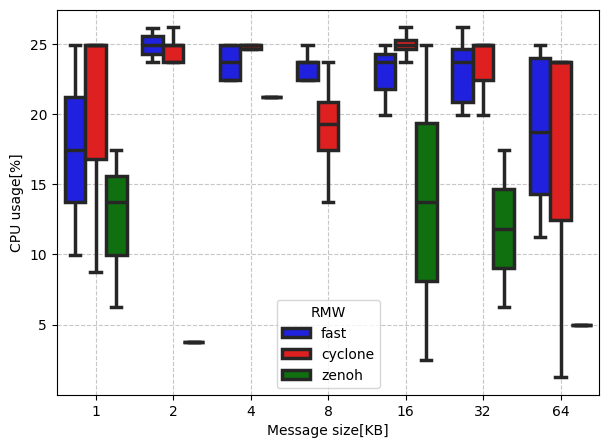}
        \caption{CPU usage in percentage on Leo02}
        \label{fig:leo02_CPU}
    \end{subfigure}
    \hfill  
    \begin{subfigure}[b]{0.49\textwidth}
        \includegraphics[width=\textwidth]{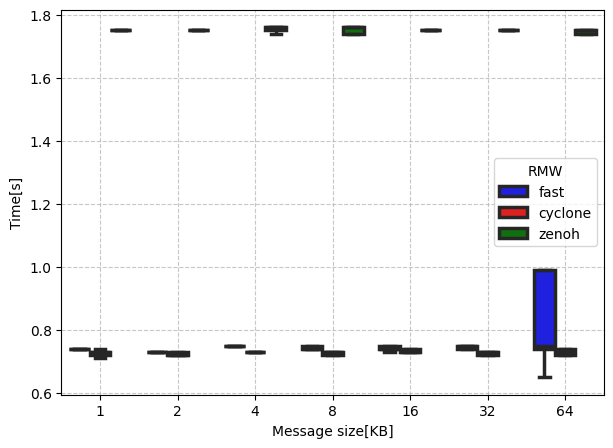}
        \caption{CPU time in second on Leo02}
        \label{fig:leo02_cpu_time}
    \end{subfigure}
    
    \caption{Box Plot representing the CPU usage in percentage}
    \label{fig:CPU}
\end{figure*}

\textcolor{red}{Figure \ref{fig:leo02_CPU} illustrates the CPU usage on the Raspberry of leo02 when sending messages of fixed sizes for both scenarios, while Figure \ref{fig:leo02_cpu_time} displays the CPU time.}
\textcolor{red}{Comparing both metrics shows interesting results, while Zenoh uses in median around two times less the CPU, it seems to use two times longer. Showing a behaviour that tends to be more lightweight for the embedded computers while being slightly slower to compute}

\begin{figure*}[htbp]
        \begin{subfigure}[b]{0.49\textwidth}
            \includegraphics[width=\textwidth]{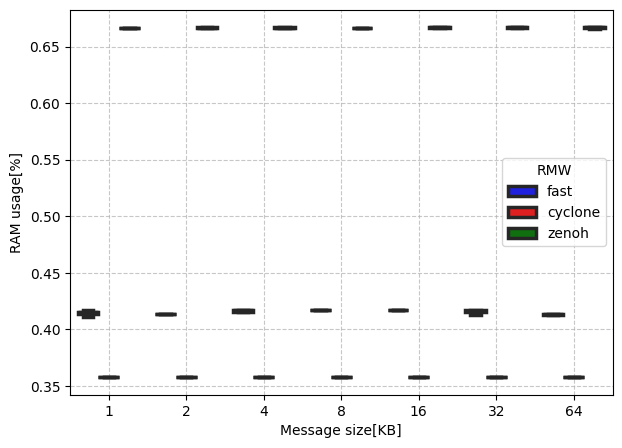}
            \caption{RAM usage in percentage on Leo02}
            \label{fig:leo02_RAM}
        \end{subfigure}
        \hfill  
        \begin{subfigure}[b]{0.49\textwidth}
            \includegraphics[width=\textwidth]{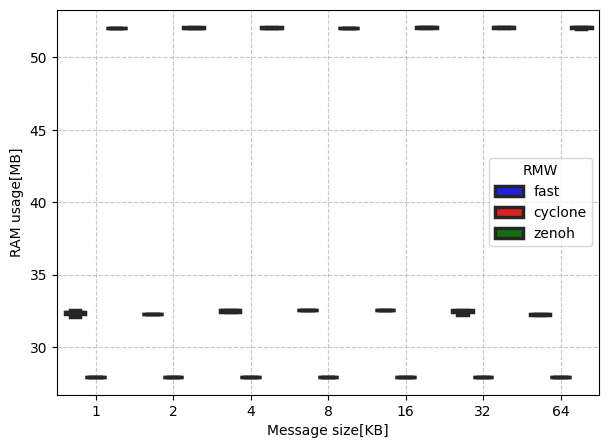}
            \caption{RAM usage in $10^7$ bytes on Leo02}
            \label{fig:leo02_ram_info}
        \end{subfigure}
    \caption{Box Plot representing the RAM usage in percentage during the scenario}
    \label{fig:RAM}
\end{figure*}

\textcolor{red}{It is notable that Zenoh uses a router that adds overhead to the overall resource consumption. However, even with this overhead, Zenoh depicts very good performances compared to the other RMW.} \textcolor{red}{While CPU consumption has a direct impact on the battery life and other algorithms' performance, RAM (Random access memory) is required to save data while computing. To this extent, Figure \ref{fig:leo02_RAM} displays the RAM percentage usage of ROS2 processes (i.e. RMW consumption) while Figure \ref{fig:leo02_ram_info} the bytes consumption of such process.}
\textcolor{red}{Both diagrams show the same data but use different representations (one as a percentage, the other one in bytes) to get the RAM consumption's magnitude fully. Those diagram have very narrow box plots but depict a constant behavior for each RMW. Starting from left: Fast, Cyclone and Zenoh shows the same amount of used memory no matter the size of the transmitted data which is really surprising for us.
Cyclone and Fast use the least amount of memory with respectively 29.3MB and 34MB representing less than 0.45\% of the 4GB of RAM of the Raspberry. On Zenoh's side, this number grows up to 54.6MB representing 0.66\%.
Zenoh takes more RAM than the other RMW, however it relies on a router to be functional. This router might add overhead to the initial 'cost' of sending a message on the network and therefore increase RAM consumption as it might have increased CPU consumption.}

\subsection{Summary}

Table \ref{tab:sumup_symbols} displays an overview of the performances of each middleware over different performance metrics. Symbols display how efficient each RMW is on a specific point of interest compared to the others. A ``++"  indicates that the system outperforms the others in the scenario, while a ``- -" indicates major performance issues compared to the other approaches. The intermediary symbols show: ``+" good results, ``+/-" acceptable results and ``-" less good results than the others. Finally, the ``?" indicates missing values. As depicted earlier, Zenoh shows outperforming result compared to the others RMW.

\begin{table}[!htbp]
\begin{tabular}{m{2.5cm}p{4cm}lllll}
                                     &                                   & Fast                       & Cyclone                           & Zenoh                                              \\ \hline
Small Messages                          & Delay                          & -                          &  \cellcolor[HTML]{FD6864} - -      & \cellcolor[HTML]{9AFF99}{\color[HTML]{333333} ++} \\
                                         & Reachability                      & -                          & -                             & \cellcolor[HTML]{9AFF99}{\color[HTML]{333333} ++} \\
                                         & Data overhead                         & -                          & +/-                           & +                                                \\
                                         & CPU Usage                         & +/-                        & +/-                           & \cellcolor[HTML]{9AFF99}{\color[HTML]{333333} ++} \\
                                         & RAM Usage                         & +/-                        & +                             & -                                                  \\
 
 \cellcolor[HTML]{EFEFEF}Medium Messages & \cellcolor[HTML]{EFEFEF}Delay     & \cellcolor[HTML]{FD6864} - - & \cellcolor[HTML]{EFEFEF} - & \cellcolor[HTML]{9AFF99}{\color[HTML]{333333} ++} \\
 \cellcolor[HTML]{EFEFEF}                & \cellcolor[HTML]{EFEFEF}Reachability & \cellcolor[HTML]{EFEFEF} - & \cellcolor[HTML]{EFEFEF} +/- & \cellcolor[HTML]{9AFF99}{\color[HTML]{333333} ++}  \\
 \cellcolor[HTML]{EFEFEF}                & \cellcolor[HTML]{EFEFEF}Data overhead & \cellcolor[HTML]{EFEFEF} - & \cellcolor[HTML]{EFEFEF} +/- & \cellcolor[HTML]{EFEFEF}{\color[HTML]{333333} -}  \\
 \cellcolor[HTML]{EFEFEF}                & \cellcolor[HTML]{EFEFEF}CPU Usage & \cellcolor[HTML]{EFEFEF} +/- & \cellcolor[HTML]{EFEFEF} +/-  & \cellcolor[HTML]{9AFF99}{\color[HTML]{333333} ++}  \\ 
 \cellcolor[HTML]{EFEFEF}                & \cellcolor[HTML]{EFEFEF}RAM Usage & \cellcolor[HTML]{EFEFEF} +/- & \cellcolor[HTML]{EFEFEF} +  & \cellcolor[HTML]{EFEFEF}{\color[HTML]{333333} -}  \\ \hline
 
 \cellcolor[HTML]{bfbfbf}Larger Messages & \cellcolor[HTML]{bfbfbf}Delay     & \cellcolor[HTML]{FD6864} - - & \cellcolor[HTML]{bfbfbf} + & \cellcolor[HTML]{bfbfbf}{\color[HTML]{333333} +} \\
 \cellcolor[HTML]{bfbfbf}                & \cellcolor[HTML]{bfbfbf}Reachability & \cellcolor[HTML]{bfbfbf} - & \cellcolor[HTML]{bfbfbf} + & \cellcolor[HTML]{bfbfbf}{\color[HTML]{333333} +}  \\
 \cellcolor[HTML]{bfbfbf}                & \cellcolor[HTML]{bfbfbf}Data overhead & \cellcolor[HTML]{bfbfbf} + & \cellcolor[HTML]{bfbfbf} +/- & \cellcolor[HTML]{bfbfbf}{\color[HTML]{333333} -}  \\
 \cellcolor[HTML]{bfbfbf}                & \cellcolor[HTML]{bfbfbf}CPU Usage & \cellcolor[HTML]{bfbfbf} +/- & \cellcolor[HTML]{bfbfbf} +/-  & \cellcolor[HTML]{9AFF99}{\color[HTML]{333333} ++}  \\ \ 
 \cellcolor[HTML]{bfbfbf}                & \cellcolor[HTML]{bfbfbf}RAM Usage & \cellcolor[HTML]{bfbfbf} +/- & \cellcolor[HTML]{bfbfbf} +  & \cellcolor[HTML]{bfbfbf}{\color[HTML]{333333} -}  \\ \hline                                           
\end{tabular}
\caption{Sumary table of RMW performances compared to each message size}
\label{tab:sumup_symbols}
\end{table}

\textcolor{red}{For small messages, Zenoh outmatches the other RMW with smaller delays, better reachability, not too high data overhead per message, and reduced CPU usage while using slightly more RAM. However, the RAM usage is still within the acceptable margin. Medium messages still depict a good picture in favour of Zenoh with mitigated data overhead and RAM usage. Large messages, on the other hand, show slightly better results but are still in favor of Zenoh. However, we recommend selecting the RMW based on the conducted experiment: if battery management and the reachability of the network are the key parameters, Zenoh might be more suited. On the other hand, if the data throughput is more important, Cyclone might be more interesting.}

\textcolor{red}{Table \ref{tab:sumup_percentage} emphasizes the difference between Zenoh and the other RMW. Out of five metrics, Zenoh surpasses the other RMW in terms of delay, reachability, and CPU usage. On the other hand, RAM usage has increased, but the order of magnitude is derisory compared to embedded computer resources nowadays. However, the results are more mitigated for the data overhead. On average, Fast offers comparable results, whereas Cyclone adds around 50\% less data overhead to transmit a ROS 2 message over the network.}

\begin{table}[!htbp]
\begin{tabular}{m{4cm}p{3.5cm}lll}
                            & Zenoh to Fast                         & Zenoh to Cyclone      \\ \hline
Delay (reduced)             & \cellcolor[HTML]{9AFF99} 76\%         & \cellcolor[HTML]{9AFF99} 69.86\% \\
Reachability (increased)    & \cellcolor[HTML]{9AFF99} 146.93\%     &  \cellcolor[HTML]{9AFF99} 58.17\% \\
Data overhead per message & \cellcolor[HTML]{FD6864} 4.36\%       & \cellcolor[HTML]{FD6864} 48.14\% \\
CPU Usage (reduced)         & \cellcolor[HTML]{9AFF99} 41.27\%      & \cellcolor[HTML]{9AFF99} 39.76\% \\
RAM Usage (increased)       & \cellcolor[HTML]{FD6864} 60.50\%      & \cellcolor[HTML]{FD6864} 86.03\%  \\
\end{tabular}
\caption{Comparison table between Zenoh and the DDSs on global performances}
\label{tab:sumup_percentage}
\end{table}

In the context of \textcolor{red}{extra-terrestrial} extreme environment exploration, \textcolor{red}{a fixed communication delay cannot be avoided, forbidding any real-time operations and control. Any network-induced delay is negligible and less interesting for this scenario.} However, data overhead, reachability and CPU usage have an high interest. 
Since the mesh network can offer limited performance (depending on the used protocol, hardware, and environment), it is designed to kick out any node that uses too much data throughput, so the selected solution should focus on optimizing the data throughput usage and, thereby, the data overhead.
Exploration rovers must execute various resource-expensive mapping and navigation software. Saving computational power should also be a concern in space-oriented scenarios. \textcolor{red}{Also, computational power directly translates into electrical power, which is a critical resource in space.}
On the other hand, all of those metrics are pointless if the nodes in the network are unreachable. This is why we choose the reachability as our key parameter regarding which RMW to choose.

\textcolor{red}{In \cite{liang_performance_2023}, the authors presents similar results, where Zenoh stands out in the DDS implementation. They also highlight Message Queuing Telemetry Transport (MQTT) as the best solution, even if there is no known implementation compatible with ROS2. It could be worth investigating an implementation with ROS. Yet, MQTT is designed for IOT and handling small messages and would probably has good performance in specific scenarios.}

\section{Conclusion and future work} \label{sec:conclusion}

This study evaluates the performances of available ROS 2 middleware (FastRTPS, Cyclone DDS and Zenoh) over a mesh network given a dynamic topology. The final choice of ROS 2 middleware will be given by the closest middleware able to fulfil the scenario: an exploration of the extreme extra-terrestrial environment using a \gls{mrs}.

The results emphasize Zenoh as a promising solution for using ROS 2 on a mesh network in an extra-terrestrial extreme environment exploration scenario for \gls*{mrs}. In this scenario, where robots have limited computing capabilities and energy while needing to maintain consistent connectivity, our focus is on power consumption and reachability, which are two key strengths of Zenoh.

However, Zenoh might not be suited for scenarios that require very low RAM usage and data throughput per message transmitted. Furthermore, Zenoh's very good results tend to be more mitigated with larger messages. Further studies on larger messages (up to one Megabyte or even larger) need to be conducted to fully picture Zenoh's performance on very large messages.

In space missions where a robot needs to reach a given position, we need to ensure the connection is maintained. Future work will focus on developing collaboration mechanisms for Connectivity maintenance in mesh networks. In order to achieve optimal networking capabilities for space multi-robot systems, future work will also focus on optimizing network resources.

\section*{Declarations}
\begin{itemize}
\item Acknowledgements: This research was funded in whole or in part by the  Luxembourg  National  Research Fund (FNR), grant references [C20/IS/14783405/FiReSpARX/Fridgen] and [CAESAR-XR/Gabriel/17679211]. For the purpose of open access and in fulfilment of the obligations arising from the grant agreement, the author has applied a Creative Commons Attribution 4.0 International (CC BY 4.0) license to any Author Accepted Manuscript version arising from this submission.
\item Funding: This research was funded in whole or in part by the  Luxembourg  National  Research Fund (FNR), grant references [C20/IS/14783405/FiReSpARX/Fridgen] and [CAESAR-XR/Gabriel/17679211]. For the purpose of open access and in fulfilment of the obligations arising from the grant agreement, the author has applied a Creative Commons Attribution 4.0 International (CC BY 4.0) license to any Author Accepted Manuscript version arising from this submission.
\item Conflict of interest/Competing interests: The authors declare no conflicts of interest.
\item Ethics approval: The authors declare that this study does not pertain to a subject that requires approval by an Ethics committee.
\item Consent to participate: The authors declare that this is not applicable to this work.
\item Consent for publication: The authors declare that this is not applicable to this work.
\item Availability of data and materials: The data are available on \url{https://github.com/Gabryss/RMW-Mesh-Experiments}, under the folder data\_analysis/data
\item Code availability: The code used to perform the experiment is available at: \url{https://github.com/Gabryss/RMW-Mesh-Experiments}. The repository also contains the code to process the data along with the appropriate documentation.
\item Authors' contributions: 
\begin{itemize}
    \item Loïck Pierre Chovet: Conceptualization, Methodology, Software, Data Analysis, Writing.
    \item Gabriel Manuel Garcia: Methodology, Software, Data Acquisition and formatting, Writing.
    \item Abhishek Bera: Methodology, Review and Editing.
    \item Antoine Richard: Methodology.
    \item Kazuya Yoshida: Review and Editing.
    \item Miguel Angel Olivares-Mendez: Review and Editing.
\end{itemize}
The authors Loïck Chovet and Gabriel Garcia contributed equally to this work.
\end{itemize}


\bibliography{sn-bibliography}

\end{document}